\documentclass{article} % For LaTeX2e
\usepackage{iclr2026_conference,times}

\usepackage{amsmath,amsfonts,bm}

\def\Secref#1{Section~\ref{#1}}
\def\eqref#1{equation~\ref{#1}}
\def\Eqref#1{Equation~\ref{#1}}
\def\1{\bm{1}}

\def\vt{{\bm{t}}}

\def\vv{{\bm{v}}}

\def\vz{{\bm{z}}}

\DeclareMathAlphabet{\mathsfit}{\encodingdefault}{\sfdefault}{m}{sl}
\SetMathAlphabet{\mathsfit}{bold}{\encodingdefault}{\sfdefault}{bx}{n}

\DeclareMathOperator*{\argmax}{arg\,max}

\usepackage{algorithm}
\usepackage{algpseudocode}
\usepackage{subcaption}

\DeclareCaptionSubType{algorithm}
\usepackage{graphicx}  %Required
\usepackage{bbm}
\usepackage{float}
\usepackage{xcolor,colortbl}
\usepackage{amsmath}
\usepackage{braket}
\usepackage{amssymb}
\usepackage{mathtools}
\usepackage{amsthm}
\usepackage{amsmath,amsfonts,bm}
\usepackage{bigstrut,bigdelim}
\usepackage{paralist}
\usepackage{diagbox}
\usepackage{wrapfig}
\usepackage{verbatim}
\usepackage{framed}
\usepackage[most]{tcolorbox}

\usepackage{subcaption}
\usepackage{multicol}
\usepackage{multirow}
\usepackage{amssymb}
\usepackage[tableposition=top]{caption}
\usepackage{hyperref}       % hyperlinks
\usepackage{cleveref}
\usepackage{acronym}
\usepackage{xspace}
\usepackage{pifont}
\usepackage{lipsum}
\usepackage{textcomp,scalerel}
\usepackage{enumitem}
\usepackage[misc]{ifsym}
\usepackage{appendix}
\usepackage{titletoc}
\usepackage{lipsum}
\usepackage[table]{xcolor}

\usepackage{siunitx}
\usepackage[table]{xcolor}

\makeatletter
\DeclareRobustCommand\onedot{\futurelet\@let@token\@onedot}
\def\@onedot{\ifx\@let@token.\else.\null\fi\xspace}

\makeatother

\makeatletter
\renewcommand{\paragraph}{%
  \@startsection{paragraph}{4}%
  {\z@}{0ex \@plus 0ex \@minus 0ex}{-1em}%
  {\hskip\parindent\normalfont\normalsize\bfseries}%
}
\makeatother

\makeatletter
\newcommand{\thickhline}{%
    \noalign {\ifnum 0=`}\fi \hrule height 1pt
    \futurelet \reserved@a \@xhline
}
\makeatother

\newcommand{\cmark}{\ding{51}}%
\newcommand{\xmark}{\ding{55}}%

\theoremstyle{plain}

\theoremstyle{definition}

\theoremstyle{remark}

\usepackage{hyperref}
\usepackage{url}
\hypersetup{hidelinks} % simplest

\usepackage{fontawesome5}

\usepackage{booktabs}
\usepackage{adjustbox}
\usepackage{xcolor}

\usepackage[utf8]{inputenc}
\usepackage{textcomp}  % Required for encoding \textbigcircle
\usepackage{scalerel}  % Required for emoji \scalerel
\def\rewardtop{\scalerel*{\includegraphics{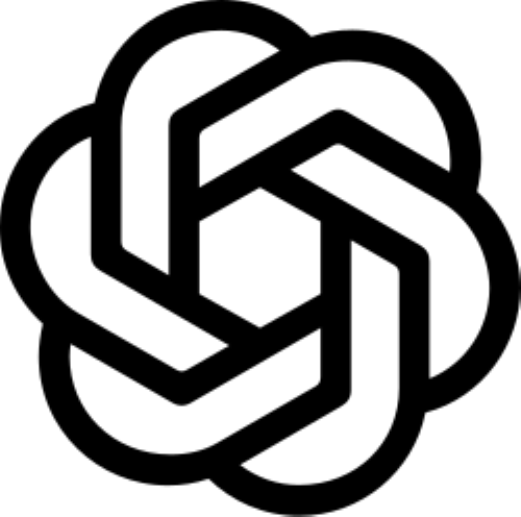}}{\textrm{\textbigcircle}}\xspace}
\def\rewardmix{\scalerel*{\includegraphics{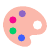}}{\textrm{\textbigcircle}}\xspace}
\def\rewardself{\scalerel*{\includegraphics{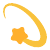}}{\textrm{\textbigcircle}}\xspace}
\def\rewardgold{\scalerel*{\includegraphics{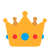}}{\textrm{\textbigcircle}}\xspace}
\def\rewardunify{\scalerel*{\includegraphics{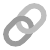}}{\textrm{\textbigcircle}}\xspace}

\newcommand{\model}{\textsc{MILR}\xspace}
\newcommand{\fulltitle}{Improving \textcolor{orange}{M}ultimodal \textcolor{orange}{I}mage Generation via Test-Time \textcolor{orange}{L}atent \textcolor{orange}{R}easoning\xspace}

\def\vv{\mathbf{v}}
\def\vt{\mathbf{t}}
\def\vz{\mathbf{z}}

\title{\model: \fulltitle}
\iclrfinalcopy
\author{
\begin{tabular}{@{}l@{}}
\textbf{Yapeng Mi}$^{1,2}$ \quad
\textbf{Yanpeng Zhao}$^{2}  \, ^{\dagger} \, \textsuperscript{\Letter}$ \quad
\textbf{Hengli Li}$^{2,3}$ \quad
\textbf{Chenxi Li}$^{2,4}$ \\
\textbf{Huimin Wu}$^{2}$ \quad
\textbf{Xiaojian Ma}$^{2}$ \quad
\textbf{Song-Chun Zhu}$^{2,3}$ \quad
\textbf{Ying Nian Wu}$^{5}$ \quad
\textbf{Qing Li}$^{2}\,\textsuperscript{\Letter}$
\end{tabular}
\\[0.35em]
\begin{tabular}{@{}l@{}}
$^{1}$University of Science and Technology of China \\
$^{2}$State Key Laboratory of General Artificial Intelligence, BIGAI \\
$^{3}$Peking University \quad
$^{4}$Tsinghua University \quad
$^{5}$University of California, Los Angeles
\end{tabular}
\\[1.0em]
\faGlobe\ \textbf{Project:}~\url{https://spatigen.github.io/milr.io/}
\quad
\faGithub\ \textbf{Code:}~\url{https://github.com/spatigen/milr}
}

\newcommand{\blfootnote}[1]{%
  \begingroup
  \renewcommand\thefootnote{}\footnote{#1}%
  \addtocounter{footnote}{-1}%
  \endgroup
}

\begin{document}

\maketitle 

\blfootnote{
\begin{tabular}{@{}l@{}}
Contact: \texttt{miyapeng78@gmail.com, yannzhao.ed@gmail.com, dylan.liqing@gmail.com}. \\
$\dagger$: Project Lead. \ \Letter: Corresponding Author.
\end{tabular}
}

\begin{abstract}
Reasoning-augmented machine learning systems have shown improved performance in various domains, including image generation. However, existing reasoning-based methods for image generation either restrict reasoning to a single modality (image or text) or rely on high-quality reasoning data for fine-tuning. To tackle these limitations, we propose \model, a test-time method that jointly reasons over image and text in a unified latent vector space. Reasoning in \model is performed by searching through vector representations of discrete image and text tokens.
Practically, this is implemented via the policy gradient method, guided by an image quality critic.
We instantiate \model within the unified multimodal understanding and generation (MUG) framework that natively supports language reasoning before image synthesis and thus facilitates cross-modal reasoning. The intermediate model outputs, which are to be optimized, serve as the unified latent space, enabling \model to operate entirely at test time. We evaluate \model on GenEval, T2I-CompBench, and WISE; it achieves state-of-the-art results on all benchmarks. Notably, on knowledge-intensive WISE, \model attains an overall score of 0.63, improving over the baseline by 80\%. Our further analysis indicates that joint reasoning in the unified latent space is the key to its strong performance. Moreover, our qualitative studies reveal \model's nontrivial ability in temporal and cultural reasoning, highlighting the efficacy of our reasoning method. 
\end{abstract}

\begin{figure}[h]
  \centering
  \includegraphics[width=1.0\linewidth]{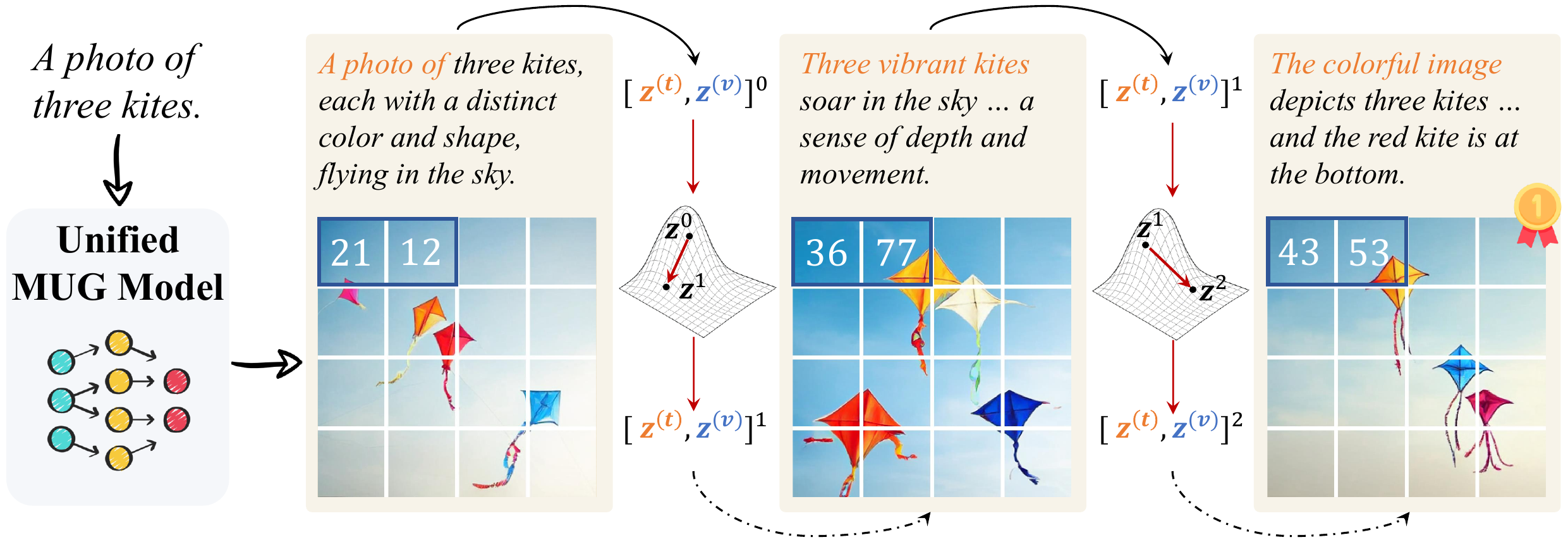}
  \caption{
  Latent reasoning of \model. The black \emph{solid} line denotes extracting the output vector representations $\vz^{k}$ of the text tokens $\vz^{(t)}$ and image tokens $\vz^{(v)}$ to be optimized, and the black \emph{dashed} line denotes decoding from the optimized latent vectors $\vz^{k+1}$, where $\vz=[\vz^{(t)},\vz^{(v)}]$.
  }
  \label{fig:teaser}
  \vspace{-0em}
\end{figure}

\section{Introduction}
% \yann{
Text-guided image generation is the task of synthesizing an image conditioned on a given text instruction. In recent years, the field has witnessed transformative progress: moving from generative adversarial models~\citep{goodfellow2014gan} to autoregressive and diffusion approaches~\citep{sun2024LlamaGen,esser2024sd3-med,flux2024,chen2025janusprounifiedmultimodalunderstanding,wu2025nep}. However, traditional models are limited in generating images in a single-shot fashion and thus are unable to resolve potential defects~\citep{huang2025t2icompbench++,niu2025wise}. Inspired by the success of reasoning-augmented LLMs---such as OpenAI o1~\citep{openai2024openaio1card} and DeepSeek-R1~\citep{deepseekai2025deepseekr1incentivizingreasoningcapability}---that can reflect on and refine their thoughts, and answer accordingly, recent works have attempted to endow image generation models with reasoning ability~\citep{guo2025can,fang2025got,zhang2025reasongenr1cotautoregressiveimage,wu2025repromptreasoningaugmentedrepromptingtexttoimage,chern2025thinkinggeneratedimages}.

Reasoning-augmented image generation models typically perform reasoning in two spaces:
language and image spaces. Language reasoning primarily involves refining instructions to make them more comprehensible to the model~\citep{wu2025repromptreasoningaugmentedrepromptingtexttoimage,li2025reflectdit}, and image reasoning iterates on the generation guided by a quality metric~\citep{guo2025can,zhuo2025reflectionflow}. 
Early works implement reasoning either in the language or in the image space, lacking a mechanism for synergistic reasoning across the two spaces. 
To fill the gap, recent works resort to unified multimodal understanding and generation (MUG;~\citet{jiang2025t2i-r1,duan2025got-r1,zhang2025reasongenr1cotautoregressiveimage}), which natively supports language reasoning before generating images (referred to as multimodal image generation), facilitating cross-modal image-text reasoning. 
Despite the success, these approaches require carefully curated reasoning data and depend on model fine-tuning, rendering them complex and costly to develop in practice.

To address these limitations, we propose Multimodal Image generation via test-time Latent Reasoning, dubbed \model.
Our core idea is to reason in a unified
\emph{latent} vector space that encodes both text and images, % into a shared representation, 
starkly different from previous methods that explicitly reason over raw images and text. We build MILR upon a Transformer-based MUG model %a Transformer-based multimodal image generation model, 
and choose the unified latent space represented by the intermediate output vectors. % of a Transformer layer. % of the model.
% and choose the unified latent space spanned by the vectors input to the final modality-specific decoding heads. 
Since the shared latent space is modality-agnostic, it provides a unified view of multimodal reasoning, reducing the modality gap and improving the overall efficacy of cross-modal reasoning. % over images and text.

Reasoning with MILR involves finding the best vector representations of image and text tokens that lead to improved image quality. We implement it using the policy gradient method~\citep{williams1992simple}, where, at test time, the reward is computed by scoring the compatibility between the generated image and the given instruction. Crucially, gradients are only back-propagated to the cross-modal latent representations (i.e., the intermediate model outputs) obtained from the forward pass (see Figure~\ref{fig:teaser}), 
% (where image tokens are generated following textual reasoning tokens; see Figure xxx), 
without altering any model parameters, and thus making MILR a test-time latent reasoning method.

We evaluate MILR on three widely-used image generation benchmarks: GenEval~\citep{ghosh2023geneval}, T2I-CompBench~\citep{huang2023t2icompbench}, and WISE~\citep{niu2025wise}. 
\model established new state-of-the-art results across all benchmarks. Notably, it achieves an overall score of 0.95 on GenEval, matching the best training-based model and outperforming the best test-time-scaling method by 4.4\%. On the more challenging WISE benchmark, \model obtains an overall score of 0.63, surpassing the strongest baseline model by 16.7\%. Our further analysis reveals that the best performance of \model is driven by its ability to perform joint image-text reasoning in a unified latent space. This unique advantage also makes \model successful in instructions that involve challenging cultural and temporal reasoning.

To summarize:
\begin{itemize}[leftmargin=*]
\item We introduce \model, a test-time reasoning-augmented method that improves image generation by performing joint image-text reasoning in a unified latent space.
\item We demonstrate the effectiveness of MILR by showing that it achieves superior performance across three different benchmarks for image generation.
\item We conduct a comprehensive analysis, perform various ablation studies, and discuss potential limitations of \model.
\end{itemize}

% }

\section{Related Works}
\paragraph{Reasoning-Augmented Image Generation.}
Remarkable progress has been achieved in reasoning-augmented LLMs~\citep{wei2022chain,deepseekai2025deepseekr1incentivizingreasoningcapability}, but effectively integrating reasoning into conventional text-guided image generation models remains a challenge~\citep{betker2023dalle,sun2024LlamaGen,esser2024sd3-med,zhao2024ultraedit,chen2025janusprounifiedmultimodalunderstanding}. 
With the development of unified multimodal understanding and generation that can generate images and text in an interleaved manner, many works have explored reasoning for text-guided image generation. The predominant paradigm uses causal language modeling to unify image and text generation as next token prediction, enabling the model to make plans via chain-of-thought before generating images~\citep{fang2025got,deng2025bagel,xiao2025mindomni,cai2025IGTR,guo2025can,jiang2025t2i-r1,duan2025got-r1}. Another line of research focuses on reasoning over images via the test-time scaling strategy~\citep{li2025reflectdit,zhuo2025reflectionflow}. These works typically rely on an external critic model to provide feedback on further improvement. Unlike all these methods that explicitly perform reasoning over raw images and text, we use test-time optimization to refine the latent representations of image and text tokens, leading to a unified cross-modal latent reasoning method.

\paragraph{Reasoning in The Latent Space.} Differently from explicit reasoning via a chain of thoughts~\citep{wei2022chain}, latent reasoning refers to an implicit reasoning mechanism applied to the latent states (e.g., intermediate outputs) of the model. As in Transformer-based models, latent reasoning is typically implemented as spatial and temporal recurrences. Spatial recurrences implicitly deepen the model by iterating over the latent states between the Transformer layers~\citep{hao2024training,cheng2024compressed,zhang2025lightthinker,shen2025codi}, while temporal recurrences refine the latent states through iterations across input tokens~\citep{dao2024transformers,geiping2025scaling}. Under the recurrences is the idea of scaling up test-time computation for inference; however, this requires the recurrent modules to be pre-trained. In a similar vein to those, we iterate over the unified image-text latent states only at test time, without introducing and updating any model parameters.

\paragraph{Reinforcement Learning for Reasoning.}
Reinforcement learning (RL) has been the key to eliciting the reasoning ability of large language models~\citep{deepseekai2025deepseekr1incentivizingreasoningcapability,openai2024openaio1card}. Inspired by the success of GRPO, which is an improved RL algorithm over PPO~\citep{schulman2017ppo} and is used in DeepSeek-R1~\citep{shao2024deepseekmathgrpo}, researchers have extended it to the multimodal understanding domain, including visual question answering~\citep{huang2025visionr1,liu2025visualrft}. In image generation, RL has also been shown to be effective~\citep{jiang2025t2i-r1,tong2025delving,jiang2025co,pan2025focusdiff,duan2025got-r1,pan2025SelfReflectiveRL,liu2025flowgrpo,xue2025dancegrpo,xiao2025mindomni}, but has been used as a training-time optimization method. Unlike them, we employ a simple algorithm, REINFORCE~\citep{williams1992simple}, for test-time optimization.

% \section{Multi-Modal Image Generation via Test-time Latent Reasoning}
\section{Method}
\label{sec:multi_modal_generation}
% In this section, we elaborate on our approach, \model

\begin{figure}[t!]
  \centering
  \includegraphics[width=1.0\linewidth]{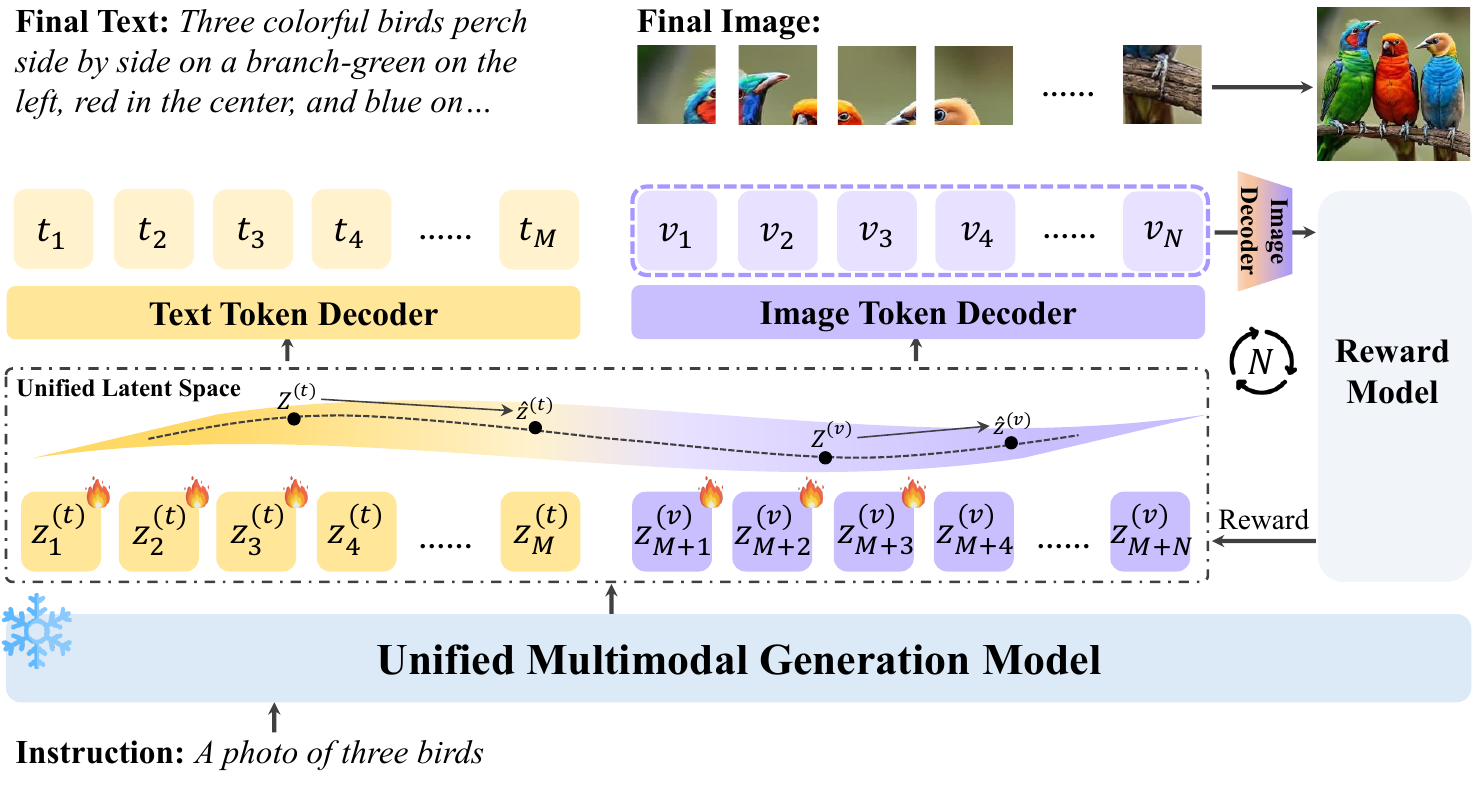}
  \vspace{-2.5em}
  \caption{Overview of \model. \model performs test-time latent reasoning in a unified latent space; it uses policy gradients to iteratively refine text \& image latents $\vz^{(t)},\vz^{(v)}$, guided by a reward model. The reward model scores each generated image conditioning on the instruction.}
  \label{fig:framework}
  \vspace{-1.5em}
\end{figure}

\subsection{Reasoning-augmented multimodal image generation}
% We refer to reasoning-augmented multimodal image generation as a unified generative framework that produces intermediate language-based reasoning tokens before synthesizing images. A predominant implementation of this framework is the unified multimodal understanding and generation model (MUG)~\cite{chen2025janusprounifiedmultimodalunderstanding,deng2025bagel}). 

We are interested in a framework for reasoning-augmented image generation that natively supports language reasoning before image synthesis.
 % because it facilitates cross-modal reasoning. 
A typical implementation is through unified multimodal understanding and generation (MUG;~\cite{chen2025janusprounifiedmultimodalunderstanding,deng2025bagel}).
Suppose the language token sequence $\vt \coloneqq t_{1:M} \coloneqq t_1, t_2, \ldots, t_M$, the image token sequence $\vv \coloneqq v_{1:N} \coloneqq v_1, v_2, \ldots, v_N$, and the given instruction $c$, MUG defines multi-modal image generation as an autoregressive generation process:
\begin{equation}
p(\vt, \vv | c) = \prod_{n=1}^{N} p(v_n | v_{1:n}, \vt, c) \prod_{m=1}^{M} p(t_m | t_{1:m}, c)\,,
\label{eq:inital}
\end{equation}
The generation of $\vv$ depends on the reasoning via language tokens $\vt$, which are generated from $c$. By analogy to textual reasoning, we refer to the generation of $\vv$ as visual reasoning. To produce the final image, $\vv$ is further passed through a pixel decoder: $V_f\sim p(\cdot|\vt,\vv,c)$.
% To produce the final image $V_f\sim p(\cdot|\vt,\vv,c)$, $\vv$ is further passed through a pixel decoder $p(\cdot|\vt,\vv,c) \coloneqq p(\cdot|\vv)$. 
% Depending on the decoder type, $\vv$ can be continuous vectors~\citep{li2024MAR,deng2025bagel} and discrete tokens~\citep{sun2024LlamaGen,chen2025janusprounifiedmultimodalunderstanding}. 
% Depending on the decoder type, $\vv$ can be continuous vectors (as is typical for latent diffusion models\citep{li2024MAR,xie2024showo,deng2025bagel}) and discrete tokens (as is usually the case for the decoder of discrete VAE~\citep{sun2024LlamaGen,chen2025janusprounifiedmultimodalunderstanding}). 
% We focus on discrete image tokens and leave the continuous version for future work, so $p(\cdot|\vt,\vv,c)$ becomes deterministic given $\vv$ in our following discussion.
Since we focus on discrete image tokens produced by a pre-trained discrete VAE~\citep{chen2025janusprounifiedmultimodalunderstanding}, the pixel decoder $p(\cdot|\vt,\vv,c)$ becomes deterministic given $\vv$. %~\footnote{Note that, given image tokens $\vv$, image decoding is independent of the language tokens $\vt$ and instruction $c$}. % in the following discussion.

% Unlike language reasoning in the form of discrete tokens, we define visual tokens as continuous vectors (with a slight abuse of notation), which are further decoded into pixel images.

Let $R(V_f, c)$ denote a reward model that scores the compatibility between the given instruction $c$ and the final image $V_f$, the goal of test-time reasoning-augmented image generation is to find an optimal pair ($\vt^{*}, \vv^{*}$) that maximizes the expected reward under $p(\cdot|\vt,\vv,c)$, without modifying any model parameters:
\begin{equation}
    \vt^{*}, \vv^{*}= \argmax_{\vt,\vv} \mathbb{E}_{V_f \sim p(\cdot|\vt,\vv,c)}[R(V_f,c)]\,.
    \label{eq: explicit_reasoning}
\end{equation}
In the case of MUG, the model itself can act as the reward function because of its multimodal understanding ability; %since its multimodal understanding capacity enables image quality assessment;
alternatively, any off-the-shelf model that has such a capability can serve the same role (see discussion in~\Secref{sec:reward-model}). Due to the infinite search space, the problem defined by~\Eqref{eq: explicit_reasoning} is generally intractable.

\subsection{Image Generation via test-time latent reasoning}\label{sec:our-model}

\subsubsection{Multi-Modal Latent Reasoning} 

Rather than searching over discrete image and text tokens, we propose searching in the unified latent vector space, that is, searching over their continuous vector representations. As in MUG, these vectors correspond to the intermediate model outputs at respective token positions. They lie in a vector space that encodes both image and text tokens, offering a unified view of visual and textual reasoning\footnote{We refer to “reasoning” as iterative updates in the latent space rather than text-based chain-of-thought.} and facilitating cross-modal reasoning (see Figure~\ref{fig:framework}).

Formally, denoting the latent representations of image and text tokens by $\vz^{(v)}=z^{(v)}_{1:N}$ and $\vz^{(t)}=z^{(t)}_{1:M}$, respectively, where $z^{(v)}, z^{(t)}\in\mathbb{R}^{d}$ are the outputs from the same Transformer layer and thus lie in a shared $d$-dimensional vector space, we can rewrite~\Eqref{eq: explicit_reasoning} as:
\begin{equation}\label{eq:implicit-reasoning}
\vz^* = \argmax_{\vz} \mathbb{E}_{V_f\sim p(\cdot|\vz.c)} [R(V_f, c)]\,, 
\end{equation}
where $\vz = [\vz^{(t)};\vz^{(v)}]$ indicates the multimodal latent representation of the token sequence $[\vt,\vv]$. We refer to this optimization problem as multimodal latent reasoning. 
% and dub our method \model, short for multimodal image generation via latent reasoning.

Given the optimal $\vz^{\star}$ from a specific model layer, to produce the final $V_f$, we need to continue the forward pass until it is decoded into discrete tokens $[\vt,\vv]$. Thus, the pixel image generation becomes:
\begin{equation}
p(V_f|\vz^*, c) = p(V_f | \vt, \vv, c) p(\vt, \vv|\vz^*)\,,
\label{eq:sampling_the_final_answer}
\end{equation}
where $p(\vt, \vv|\vz^*)$ represents the remaining forward pass of MUG starting with $\vz^*$.

\subsubsection{Gradient-based optimization for latent reasoning}

In general, the problem defined by~\Eqref{eq:implicit-reasoning} does not admit a closed-form solution, so we resort to REINFORCE~\citep{williams1992simple}, a policy gradient optimization method. We note that it has been applied to fully textual reasoning in language tasks~\citep{li2025seekdarkreasoningtesttime}, but we, for the first time, extend it to unified multimodal latent reasoning for image generation.

With REINFORCE, we can formulate the cross-modal optimization process as:
\begin{flalign}
\vz^{k+1} &\leftarrow \vz^{k} + \eta \cdot \mathcal{J}(\vz^k), 
\\
\mathcal{J}(\vz^{k}) &= 
        \mathbb{E}_{V_f\sim p(\cdot|\vz^{k}, c)}[R(V_f, c) \nabla_{\vz}\log( p(\vt, \vv|\vz^{k}))]\,,
\label{eq:milr-optimization}  
\end{flalign}

where $\eta$ is the learning rate. For efficiency, we choose as $\vz$ the outputs of the last Transformer layer, i.e., the inputs to the final modal-specific decoding heads. Moreover, we approximate gradients $\mathcal{J}(\vz)$ using a single sampled pair $(\vt, \vv)$. The gradients are back-propagated only to the model outputs $\vz$, without altering any model parameters, and thus making \model a test-time reasoning method.

Naively, we would optimize\footnote{We experimented with different optimization strategies, see details in \Cref{appendix:algorithm}.} all $M+N$ latents in $z_{1: M+N}$, but searching using only the guidance of a reward model is potentially biased, and it does not leverage the generative capacity of MUG for better exploration; instead, we optimize only the first $\lambda_{t}M$ (where $\lambda_{t}\in (0, 1]$) latents for text. After decoding them into discrete tokens, we complete textual reasoning via the standard autoregressive generation conditioned on them. 
% As shown by~\cite{li2025seekdarkreasoningtesttime}, 
As we will later see, this simple strategy strikes a good balance between efficiency and performance (see~\Secref{sec:analysis-hyperparam}). 
% in fully textual reasoning, and we will later see that it extends to our multimodal reasoning setting. 
For visual reasoning, we adopt a similar strategy and optimize the first $\lambda_{v}N$ (where $\lambda_{v}\in(0, 1]$) latents.\footnote{In each iteration, the text token number $M$ may vary while the image token number $N$ remains unchanged.} This is further supported by the observations of~\cite{hu2025anchortokenmatchingimplicit}: the first few tokens govern the global structure of the image, while the remaining tokens primarily influence high-frequency details.\footnote{Rather than optimizing a prefix of image token sequence, we can optimize a random subset of image tokens, but this strategy proves worse than prefix optimization (see Table~\ref{benchmark:random_image} in~\Cref{appendix:algorithm}).} 
% \myp{The full algorithm is shown in \Cref{appendix:algorithm}.}

% Due to the dependency of visual reasoning on textual reasoning, the optimization problem is conceptually similar to \yann{or equivalent to?@henry} bilevel optimization \yann{cite}; therefore, part from the default setup of

% \textbf{Algorithm Variants}:
% By default, \model jointly optimizes $\vz^{(t)}$ and $\vz^{(v)}$ in each gradient descent step. As an alternative, we can optimize $\vz^{(t)}$ and $\vz^{(v)}$ in an alternating fashion (denoted by \model-Alt), akin to the strategy used in coordinate descent.
% % a strategy widely used for bilevel optimization problems. 
% We further propose another variant of \model that first optimizes $\vz^{(t)}$ to its approximate optimum, followed by optimizing $\vz^{(v)}$ to its approximate optimum (denoted by \model-T2V). We illustrate the three algorithms in~\Cref{appendix:algorithm}.

\section{Experiments}
\subsection{Experimental Settings}
\label{sec:experimental_settings}
% \textbf{Benchmarks.} 
% We conduct experiments on three benchmarks widely used in image generation: GenEval~\citep{ghosh2023geneval}, T2I-CompBench~\citep{huang2023t2icompbench}, and WISE~\citep{niu2025wise}. 
% GenEval is an object-centric benchmark that evaluates compositional properties, such as position, count, and color. T2I-CompBench assesses compositional image generation with a special focus on attribute binding, object relationships, generative numeracy, and complex compositions. WISE challenges image generation models on their understanding of world knowledge. Collectively, they provide comprehensive evaluations of \model.

\textbf{Benchmarks and Baselines.} We conduct experiments on three benchmarks widely used in image generation: GenEval~\citep{ghosh2023geneval}, T2I-CompBench~\citep{huang2023t2icompbench}, and WISE~\citep{niu2025wise}. We compare \model against the following three sets of baselines:
% non-reasoning models, training-based reasoning models, and test-time reasoning models.
\begin{itemize}[leftmargin=*]%, topsep=0pt, noitemsep]
% \item \textbf{Foundation Image Generation Models.}
\item \textbf{Non-reasoning models} refer to the classical image generation models that synthesize images from a given instruction in a single shot, without refining the instruction. We consider diffusion models such as FLUX.1-dev~\citep{flux2024}, DALL$\cdot$E~3~\citep{betker2023dalle}, and SD3-Medium~\citep{esser2024sd3-med}, autoregressive models such as LlamaGen~\citep{sun2024LlamaGen} and Emu3~\citep{wang2024emu3}, and hybrid autogressive-diffusion models such as BAGEL~\citep{deng2025bagel} and GPT-4o~\citep{openai_4o_image_generation}.

% \item \textbf{Training-based Methods.}
\item \textbf{Training-based reasoning models} refer to the image generation models that acquire the reasoning ability through training, including
% category comprises reinforcement-learning (RL) fine-tuning with parameter updates, including 
GoT-R1~\citep{duan2025got-r1}, T2I-R1~\citep{jiang2025t2i-r1}, Flow-GRPO~\citep{liu2025flowgrpo}, and GRPO- and DPO-tuned Janus-Pro~\citep{tong2025delving}.

% \item \textbf{Inference-time Methods.}
\item \textbf{Test-time reasoning models} refer to the models that admit reasoning through tailored inference strategies, including 
% applied during test-time, including 
Reflect-DiT~\citep{li2025reflectdit} and ReflectionFlow~\citep{zhuo2025reflectionflow}, which uses language feedback, and Best-of-N and PARM~\citep{guo2025can}, which rely on search. % and ranking.% text-enhanced reasoning, and search-based approaches (Best-of-N, PARM~\citep{guo2025can}).

\end{itemize}

\textbf{Hyperparameters and configurations.}
We choose Janus-Pro~\citep{chen2025janusprounifiedmultimodalunderstanding}, an autoregressive model, as the MUG model. 
% as the unified multimodal understanding and generation model.
% an autoregressive unified model that supports joint multimodal modeling, enabling both understanding and generation. 
For the portions of text and image tokens that are optimized, we perform grid search on a validation split sampled from GenEval and empirically set $\lambda_{t}=0.2$ for text and $\lambda_{v}=0.02$ for images (see Figure~\ref{fig:text_image_analysis}). We use the Adam optimizer~\citep{kingma2014adam}, where the learning rate is empirically set to 0.03. For each benchmark, we use its own evaluation toolkit as the reward model, 
% following the best practices of
following previous work~\citep{liu2025flowgrpo,jiang2025t2i-r1,tong2025delving}.
We conduct all experiments on a single NVIDIA A100 80GB GPU. A discussion on model efficiency can be found in~\Cref{app:efficiency}.

\definecolor{bestbg}{HTML}{FFF2CC}    % 柔和浅米黄（best）
\definecolor{secondbg}{HTML}{E6F2FF}  % 柔和浅蓝（second best）
\definecolor{milrrow}{HTML}{F7F3E8}   % 卡片米色（我们的行）
\definecolor{groupbg}{gray}{0.95}     % 分组标题底色

% 语义包装：最好/次优
\newcommand{\best}[1]{\cellcolor{bestbg}\textbf{#1}}
\newcommand{\secondbest}[1]{\cellcolor{secondbg}\underline{#1}}

% 分组标题行（比 \cmidrule 更柔和）
\newcommand{\groupheader}[1]{\rowcolor{groupbg}\multicolumn{8}{c}{\textit{#1}}\\}

\begin{table}[t!]
% \begin{table}[H]
% \vspace{-1.5em}
\centering
\caption{Results on GenEval.
The best score is in \text{bold} and the second best is \text{underlined}. }
\label{benchmark:Geneval}
\begin{adjustbox}{width=\linewidth}
\begin{tabular}
% {l@{\hspace{0.5cm}}ccccccc}
{lccccccc}
\toprule
\multicolumn{1}{c}{\bf Method} &
{\bf Single Obj. $\uparrow$} &
{\bf Two Obj. $\uparrow$} &
{\bf Counting $\uparrow$} &
{\bf Colors $\uparrow$} &
{\bf Position $\uparrow$} &
{\bf Attr. Binding $\uparrow$} &
{\bf Overall $\uparrow$}
\\
\cmidrule{1-8}
% \rowcolor{groupbg}
\multicolumn{8}{c}{\textit{Non-reasoning Models}}\\
\cmidrule{1-8}
LlamaGen~\citep{sun2024LlamaGen}  & 0.71 & 0.34 & 0.21 & 0.58 & 0.07 & 0.04 & 0.32 \\
Emu3~\citep{wang2024emu3}      & 0.98 & 0.71 & 0.34 & 0.81 & 0.17 & 0.21 & 0.54 \\
FLUX.1-dev~\citep{flux2024}      & 0.98 & 0.79 & 0.73 & 0.77 & 0.22 & 0.45 & 0.66\\
DALL-E 3~\citep{betker2023dalle}     & 0.96 & 0.87 & 0.47 & 0.83 & 0.43 & 0.45 & 0.67 \\
% Show-o~\citep{xie2024showo}    & 0.98 & 0.80 & 0.66 & 0.84 & 0.31 & 0.50 & 0.68 \\
SD3-Medium~\citep{esser2024sd3-med}             & 0.99 & 0.94 & 0.72 & 0.89 & 0.33 & 0.60 & 0.74 \\
BAGEL~\citep{deng2025bagel} & 0.99 & 0.94 & 0.81 & 0.88 & 0.64 & 0.63 & 0.82 \\
GPT-4o~\citep{openai_4o_image_generation}      & 0.99 & 0.92 & 0.85 & 0.91 & 0.75 & 0.66 & 0.85 \\

\cmidrule{1-8}
% \rowcolor{groupbg}
\multicolumn{8}{c}{\textit{Training-based Reasoning Models}}\\
\cmidrule{1-8}
GoT-R1~\citep{duan2025got-r1} & 0.99 & 0.94 & 0.50 & 0.90 & 0.46 & 0.68 & 0.75 \\
T2I-R1~\citep{jiang2025t2i-r1} & 0.99 & 0.91 & 0.53 & 0.91 & 0.76 & 0.65 & 0.79 \\
Flow-GRPO~\citep{liu2025flowgrpo} & \textbf{1.00} & \textbf{0.99} & \textbf{0.95} & 0.92 & \textbf{0.99} & \underline{0.86} & \textbf{0.95}   \\
ReasonGen-R1~\citep{zhang2025reasongenr1cotautoregressiveimage}  & 0.99 & 0.94 & 0.62 & 0.90 & 0.84 & 0.84 & 0.86\\
Janus-Pro-7B(+GRPO)~\citep{tong2025delving} & 0.99 & 0.87 & 0.61 & 0.87 & 0.82 & 0.68 & 0.81\\
Janus-Pro-7B(+DPO)~\citep{guo2025can}      & 0.99 & 0.89 & 0.65 & 0.92 & 0.82 & 0.72 & 0.83\\

\cmidrule{1-8}
% \rowcolor{groupbg}
\multicolumn{8}{c}{\textit{Test-time Reasoning Models}}\\
\cmidrule{1-8}
Reflect-DiT~\citep{li2025reflectdit} & 0.98 & 0.96 & 0.80 & 0.88 & 0.66 & 0.60 & 0.81\\
ReflectionFlow~\citep{zhuo2025reflectionflow} & \textbf{1.00} & \underline{0.98} & \underline{0.90} & \underline{0.96} & 0.93 & 0.72 & \underline{0.91} \\
Janus-Pro-7B(+Text Enhanced Reasoning)      & 0.98 & 0.91 & 0.55 & 0.89 & 0.74 & 0.67 & 0.79\\
Janus-Pro-7B(+Best-of-N)      & 0.99 & 0.96 & 0.89 & 0.93 & 0.92 & 0.80 & \underline{0.91}\\
Janus-Pro-7B(+PARM)~\citep{guo2025can}      & \textbf{1.00} & 0.95 & 0.80 & 0.93 & 0.91 & 0.85 & \underline{0.91}\\
\cmidrule{1-8}
Janus-Pro-1B~\citep{chen2025janusprounifiedmultimodalunderstanding}      & 0.98 & 0.82 & 0.51 & 0.89 & 0.65 & 0.56 & 0.73 \\
\rowcolor{groupbg}
\textbf{Janus-Pro-1B+\model} & \bfseries 1.00 & 0.91 & 0.78 & 0.92 & 0.86 & \underline{0.86} & 0.89\\
Janus-Pro 7B~\citep{chen2025janusprounifiedmultimodalunderstanding}      & 0.98 & 0.85 & 0.56 & 0.89 & 0.77 & 0.64 & 0.78\\
\rowcolor{groupbg}
\textbf{Janus-Pro-7B+\model} & \textbf{1.00} & 0.96 & \underline{0.90} & \textbf{0.98} & \underline{0.98} & \textbf{0.91} & \textbf{0.95}\\

\bottomrule
\end{tabular}
\end{adjustbox}
\vspace{-0.6em}
\end{table}

% \begin{table}[tp]
\begin{table}[t!]
\centering
\caption{Results on T2I-CompBench and WISE. The best is in \text{bold}, and the second is in \text{underlined}.
% The best score is in \text{bold}, and the second best is \text{underlined}.
}
\label{benchmark:T2IandWise}
\begin{adjustbox}{width=\linewidth}
% 列：Method + GenEval(7列) + T2I-CompBench(1列 overall) + WISE(1列)
\begin{tabular}{l *{7}{c} c}
\toprule
\multicolumn{1}{c}
{\multirow{2}{*}{\bf Method}} & \multicolumn{7}{c}{\bf T2I-CompBench} & \multicolumn{1}{c}{\bf WISE} \\
\cmidrule(lr){2-8}\cmidrule(lr){9-9}
% \multicolumn{1}{c}{\bf Method} &
&
{Color $\uparrow$} &
{Shape $\uparrow$} &
{Texture $\uparrow$} &
{Spatial $\uparrow$} &
{Non-Spatial $\uparrow$} &
{Complex. $\uparrow$} &
{Overall $\uparrow$} &
{Avg $\uparrow$} \\
\midrule

\multicolumn{9}{c}{\textit{Non-reasoning Models}}\\
\cmidrule(lr){1-9}
PixArt-$\alpha$~\cite{chen2023pixart} & {0.6690} & 0.4927 & {0.6477} & 0.2064 & \underline{0.3197} & 0.3433 & 0.4465 & 0.47\\
FLUX.1-dev~\cite{flux2024}  & 0.7407 & 0.5718 & 0.6922 & 0.2863 & 0.3127 & 0.3703 & 0.4957 & 0.50\\
DALL-E 3~\cite{betker2023dalle} & 0.7785 & \textbf{0.6209} & 0.7036 & 0.2865 & 0.3003 & 0.3773 & 0.5112 & -\\
SD3-Medium~\cite{esser2024sd3-med}  & 0.8132 & \underline{0.5885} & \underline{0.7334} & 0.3200 & 0.3140 & 0.3771 & 0.5244 & 0.42\\
Show-o~\cite{xie2024showo} & 0.5600 &0.4100 &0.4600 &0.2000 &0.3000 & 0.2900 & 0.3700 & 0.30\\
BAGEL~\cite{deng2025bagel} & 0.8027 &0.5685 &0.7021 &0.3488 &0.3101 & 0.3824 & 0.5191 & 0.52\\
\cmidrule{1-9}
\multicolumn{9}{c}{\textit{Training-based Reasoning Models}}\\
\cmidrule{1-9}
T2I-R1~\cite{jiang2025t2i-r1} & 0.8130 & 0.5852 & 0.7243 & 0.3378 & 0.3090 & \underline{0.3993} & \underline{0.5281} & \underline{0.54}\\
GoT-R1~\cite{duan2025got-r1} & \underline{0.8139} & 0.5549 & \textbf{0.7339} & 0.3306 & 0.3169 & \textbf{0.3944} & 0.5241 & -\\
Janus-Pro-7B(+GRPO)~\cite{tong2025delving} & 0.7721 & 0.5366 & 0.7317 & 0.2869 & 0.3087 & 0.3697 & 0.5010 & -\\
% Janus-Pro-7B(+DPO)~\cite{tong2025delving} & 0.8783 & 0.6591 & 0.7919 & 0.4076 & 0.3153 & 0.4164 & 0.5781 & -\\
\cmidrule{1-9}
\multicolumn{9}{c}{\textit{Test-time Reasoning Models}}\\
\cmidrule{1-9}
Show-o + PARM~\cite{guo2025can} & 0.7500 & 0.5600 & 0.6600 & 0.2900 & 0.3100 & 0.3700 & 0.4900 & -\\
Janus-Pro-7B(+Text Enhanced Reasoning)      & 0.7087 & 0.4419 & 0.5821 & 0.2597 & 0.3072 & 0.3761 & 0.4459 & 0.46\\
Janus-Pro-7B(+Best-of-N)      & 0.7089 & 0.4925 & 0.7089 & \underline{0.3542} & \textbf{0.3262} & 0.3721 & 0.4938 & 0.52\\

\cmidrule{1-9}
Janus-Pro-1B~\cite{chen2025janusprounifiedmultimodalunderstanding} & 0.3411 & 0.2261 & 0.2696 & 0.0968 & 0.2808 & 0.2721 & 0.2478 & 0.26\\
\rowcolor{groupbg}
\textbf{Janus-Pro-1B+\model} & 0.6066 & 0.2796 & 0.4177 & 0.2796 & 0.2622 & 0.2613 & 0.3512 & 0.40\\
Janus-Pro-7B~\cite{chen2025janusprounifiedmultimodalunderstanding} & 0.6359 & 0.3528 & 0.4936 & 0.2061 & 0.3085 & 0.3559 & 0.3921 & 0.35\\
\rowcolor{groupbg}
\textbf{Janus-Pro-7B+\model} & \textbf{0.8508} & 0.5117 & 0.6949 & \textbf{0.4613} & 0.3078 & 0.3684 & \textbf{0.5325} & \textbf{0.63}\\

\bottomrule
\end{tabular}
\end{adjustbox}
\vspace{-1em}
\end{table}

\subsection{Main Results}\label{sec:main-results}
\model achieves state-of-the-art results on GenEval, one of the most widely-used benchmarks for image generation (see Table~\ref{benchmark:Geneval}). It improves over the base Janus-Pro-7B by 0.17, with the largest increases obtained from \textit{Counting} (+0.34), \textit{Position} (+0.21), and \textit{Attribute Binding} (+0.27). Notably, \model surpasses frontier non-reasoning models such as SD3-Medium, BAGEL, and GPT-4o (+12\%). Compared with training-based reasoning models (e.g., GoT-R1 and T2I-R1), \model performs better and requires no parameter tuning. For fairness, we also compare \model with test-time reasoning models. Surprisingly, it outperforms ReflectionFlow and PARM (+4.5\%) that rely on scaling up test-time computation, demonstrating the superiority of our test-time optimization method.

We further evaluate \model on two additional benchmarks: T2I-CompBench and WISE (see Table~\ref{benchmark:T2IandWise}). Again, it achieves the best performance on both, highlighting the robustness of our model. 
Specifically, on T2I-CompBench, \model improves over the base Janus-Pro-7B by a large margin (+0.14) and slightly outperforms T2I-R1, a strong training-based reasoning model. 
On WISE, which emphasizes world knowledge understanding, \model outperforms the base Janus-Pro-7B (+80\%) and the second-best model T2I-R1 (+16.7\%), implying the importance of reasoning in comprehending knowledge-intensive instructions (see our qualitative studies of reasoning trajectories in Figure~\ref{fig:case_study}).

\begin{figure}[t!]
  \centering
  \includegraphics[width=1.0\linewidth]{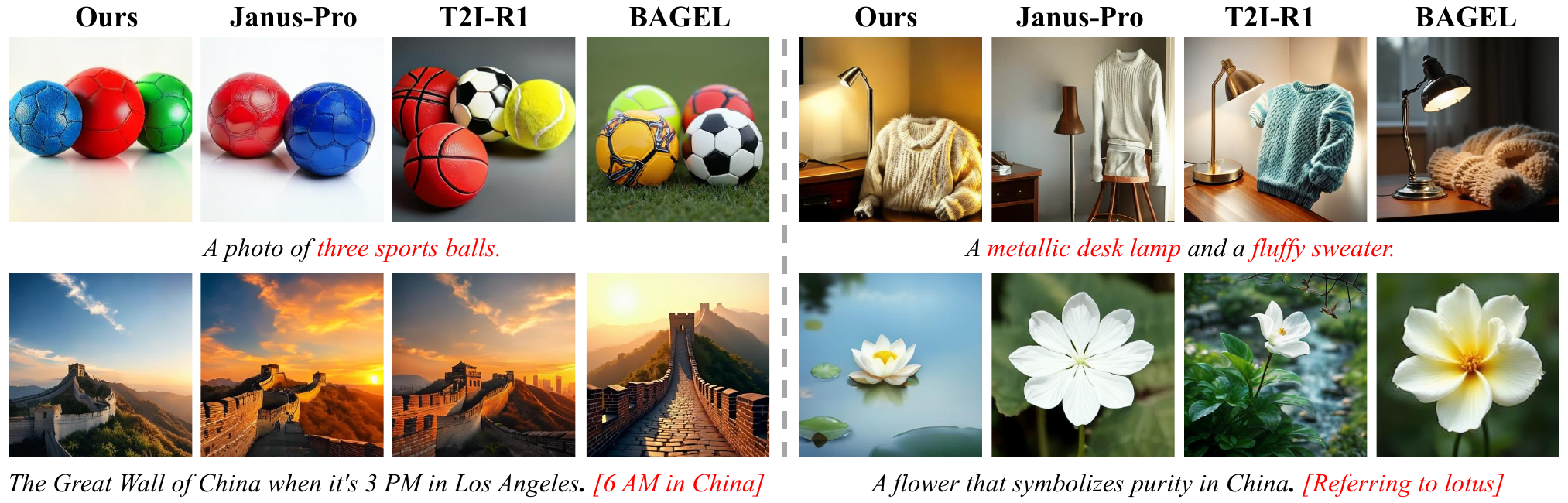}
  \caption{Qualitative studies on theree benchmarks. Reasoning cues are highlighted in red.}
  \label{fig:case_study_compare}
  \vspace{-0.25em}
\end{figure}

\paragraph{\model is capable of geometric, temporal,  and cultural reasoning.} Next, we conduct a qualitative study on knowledge-intensive WISE. Surprisingly, \model demonstrates nontrivial ability in geometric, temporal and cultural reasoning (see Figure~\ref{fig:case_study_compare}). Take the prompt "The Great Wall of China when it's 3 PM in Los Angeles", \model correctly infers the time difference from its geometric knowledge of China and Los Angeles, and concludes that "The Great Wall at Dawn" is of interest. As another example, MILR correctly infers that lotus symbolizes purity in Chinese culture.

% presents qualitative comparisons that highlight \model’s strengths in temporal reasoning (e.g., mapping ``3 PM in Los Angeles'' to 6 AM in China) and cultural reasoning (e.g., associating the lotus with purity in China), together with counting and attribute composition.

% \begin{figure}[t!]
%   \centering
%   \includegraphics[width=1.0\linewidth]{figures/case_iteration_v2.pdf}
%   \caption{Quality case study.}
%   \label{fig:case_iteration}
% \end{figure}

\paragraph{Joint image-text reasoning in the unified latent space leads to the best performance.} To better understand the contribution of each modality to the strong performance of \model, we individually ablate the latent optimization of images and text, denoted by ``w/o image" and ``w/o text", respectively (see Table~\ref{ablations}). First, we find that both of them exceed the base model (w/o MILR) by a large margin (e.g., $>0.21$ on WISE), and optimizing both modalities leads to the best performance. Interestingly, text-only optimization (w/o image) fares slightly better than image-only optimization (w/o text), and approaches the performance of our best model (+MILR) on all benchmarks, suggesting substantial room for improvement in the language understanding component of MUG-based image generation models.

\begin{table}[t!]
\centering
\caption{Ablations of \model on GenEval, T2I-CompBench, and WISE}
% Obj.: Object. Attr.:Attribution.
\label{ablations}
\begin{adjustbox}{width=\linewidth}
\begin{tabular}{l *{7}{c} c c}
\toprule
\multicolumn{1}{c}
{\multirow{2}{*}{\bf Method}} & \multicolumn{7}{c}{\bf GenEval} & \multicolumn{1}{c}{\bf T2I-CompBench} & \multicolumn{1}{c}{\bf WISE} \\
\cmidrule(lr){2-8}\cmidrule(lr){9-9}\cmidrule(lr){10-10}
% \multicolumn{1}{c}{\bf Method} &
&
{Single Obj.} &
{Two Obj.} &
{Counting} &
{Color} &
{Pos.} &
{Attr. Binding} &
{Overall} &
{Overall} &
{Avg} \\
\midrule
% \textbf{Janus-Pro-1B+\model} & 1.00 & 0.91 & 0.78 & 0.92 & 0.86 & 0.86 & 0.89 & 0.3512 & 0.40 \\
% \hspace{0.8em}\textit{w/o Image Latent Opt.} & 1.00 & 0.92 & 0.73 & 0.89 & 0.86 & 0.86 & 0.88 & 0.3509 & 0.38 \\
% \hspace{0.8em}\textit{w/o Text Latent Opt.} & 0.99 & 0.92 & 0.78 & 0.95 & 0.76 & 0.84 & 0.87 & 0.3422 & 0.34 \\
% \cmidrule(lr){1-10}
\text{Janus-Pro-7B+\model (ours)} & 1.00 & 0.96 & 0.90 & 0.98 & 0.98 & 0.91 & \textbf{0.95} & \textbf{0.5325} & \textbf{0.63} \\
\hspace{0.8em}\text{w/o \model} & 0.98 & 0.85 & 0.56 & 0.89 & 0.77 & 0.64 & 0.78 & 0.3921 & 0.35 \\
\hspace{0.8em}\text{w/o Image} & 1.00 & 1.00 & 0.91 & 0.95 & 0.95 & 0.88 & 0.94 & 0.5210 & 0.61 \\
\hspace{0.8em}\text{w/o Text} & 1.00 & 0.95 & 0.88 & 0.91 & 0.97 & 0.89 & 0.93 & 0.5043 & 0.56 \\
\bottomrule
\end{tabular}
\end{adjustbox}
\vspace{-1.5em}
\end{table}

\subsection{Analysis}

\subsubsection{Hyperparameters}\label{sec:analysis-hyperparam}
We analyze three important hyperparameters of \model: (1) the maximum optimization step $T$, (2) the portion of text tokens to be optimized $\lambda_{t}$ in text-only optimization, and (3) the proportion of image tokens to be optimized $\lambda_{v}$ in image-only optimization. We perform analysis 1 on all three benchmarks and analyses 2 and 3 on a held GenEval validation split. For each setting, we run \model three times with different random seeds and report the average scores.

\begin{figure}[t!]
  \centering
  \includegraphics[width=1.0\linewidth]{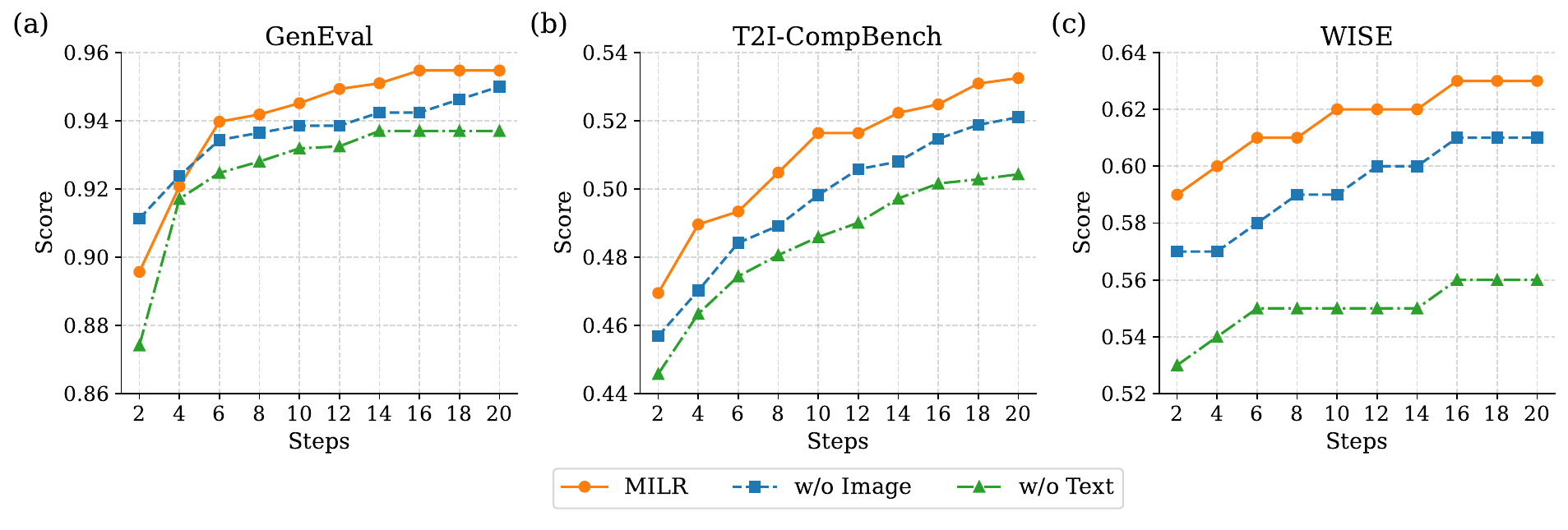}
  \caption{Performance across three benchmarks for varying optimization steps.}
  \label{fig:steps_analysis}
  \vspace{-1em}
\end{figure}

\paragraph{Scaling up the number of optimization steps improves performance on all benchmarks.} \model admits test-time compute scaling by increasing the number of optimization steps. We illustrate and compare three setups: \model, text-only optimization, and image-only optimization (see Figure~\ref{fig:steps_analysis}).
% We illustrate the performance of \model, text-only optimization, and image-only optimization with different optimization steps (see Figure~\ref{fig:steps_analysis}). 
On all benchmarks, increasing the number of optimization steps leads to consistent improvements, with the best performance achieved at step 16, after which model performance plateaus. Moreover, among the three setups, optimizing both images and text (i.e., \model) yields the best performance for almost all steps, except for a few early steps on GenEval, suggesting that \model is well-suited for test-time compute scaling.
% . We conjecture that with few steps, joint text–image optimization needs more iterations to show its edge, whereas single-modality updates better exploit the short horizon.

\begin{figure}[t!]
  \centering
  \includegraphics[width=1.0\linewidth]{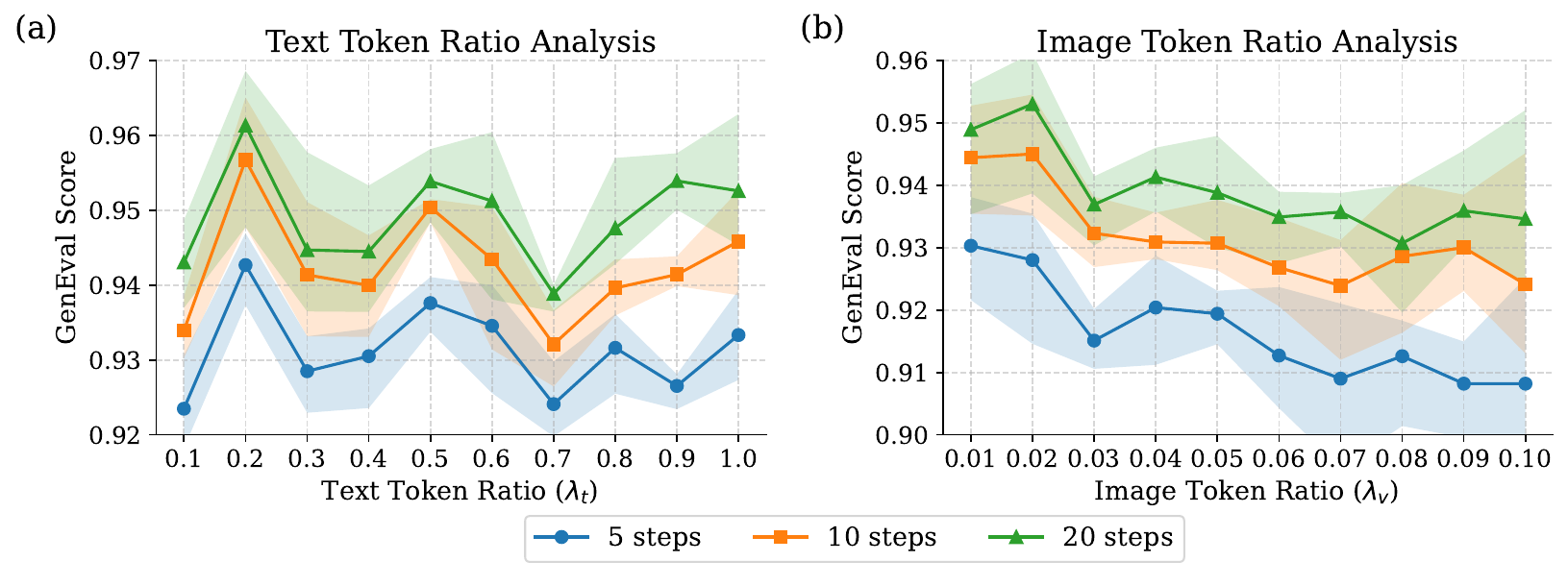}
  \caption{GenEval scores with varying optimization ratios of text and image tokens.}\label{fig:text_image_analysis}
  \vspace{-1.5em}
\end{figure}

% \noindent\textbf{Text Ratio.}
\paragraph{Optimizing a moderate amount (e.g., 20\%) of text tokens leads to the best performance.}
When varying $\lambda_{t}$ from 0.1 to 1.0, text-only optimization fluctuates between 0.91 and 0.96, and reaches the highest score at $\lambda_{t}=0.2$ (see Figure~\ref{fig:text_image_analysis}\,(a)). We observe the same trend for different numbers of optimization steps. All of these suggest that \model is relatively robust to text optimization. Moreover, we empirically find that the optimal $\lambda_{t}=0.2$ corresponds to the prompts that elicit more coherent reasoning (see examples in Figure~\ref{fig:text_cot_case}).

% \noindent\textbf{Image Ratio.}
\paragraph{Optimizing a tiny amount (e.g., 2\%) of image tokens gives rise to the best result.}
% ratio updates help; larger ratios hurt.}
\cite{hu2025anchortokenmatchingimplicit} have empirically shown that, in autoregressive image generation, early-stage tokens govern the overall image structure, and perturbing the first 20\% of image tokens results in significant structural deviations. Therefore, we conservatively adopt a very small $\lambda_{v}$, varying it from 0.01 to 0.1 (see Figure~\ref{fig:text_image_analysis}\,(b)). Surprisingly, optimizing only the first 2\% of all image tokens already yields peak performance, while further increasing $\lambda_{v}$ tends to degrade it (see examples in Figure~\ref{fig:image_k_case}).

\vspace{-0.45em}
\subsection{Reward Models}\label{sec:reward-model}
The reward model is a crucial component of \model, providing learning signals for latent reasoning. Following previous work~\citep{liu2025flowgrpo,tong2025delving}, we have used the benchmark's evaluator as the reward model (denoted by \rewardgold\textit{OracleReward}), but in real-world scenarios, oracle rewards are usually unknown, and it is difficult to design domain-specific rewards. To show that \model is effective without the reliance on \rewardgold, we test it with a set of off-the-shelf reward models on GenEval:
% we collect a diverse set of off-the-shelf reward models and configure \model with each of them individually. We consider the following reward models: 

% While prior experiments used benchmark-specific evaluators as reward models, an approach we term \rewardgold\textit{OracleReward}, this approach risks overfitting to a specific metric and may not generalize well to the unknown or changing reward models at test-time. To assess this robustness, we evaluate \model on GenEval under a diverse set of widely-used reward models:

% \yann{better to find emoji's to symbolize different critics}

% \begin{itemize}[leftmargin=*]%, topsep=0pt, noitemsep]
\begin{itemize}[leftmargin=*, topsep=0pt, itemsep=1.6pt, parsep=0pt, partopsep=0pt]
\item \rewardself \textit{SelfReward} uses MUG itself (e.g., Janus-Pro in this work) to evaluate images.
\item \rewardtop \textit{GPT-4o} represents a frontier critic for assessing image quality~\citep{hurst2024gpt4o}.
\item \rewardunify \textit{UnifiedReward} is specifically tuned for a unified evaluation of MUG~\citep{wang2025unifiedreward}.
\item \rewardmix \textit{MixedReward} is a composite critic for more comprehensive evaluation. It aggregates rewards from specialized models, including GroundingDINO~\citep{liu2024grounding} (evaluating object detection), GIT~\citep{wang2022git} (judging colors), and \rewardunify (assessing aesthetics).
% \item \textbf{OracleReward} is the standard evaluation model released along with GenEval~\citep{ghosh2023geneval}.
\end{itemize}

\begin{figure}[t!]
  \centering
  \includegraphics[width=1.0\linewidth]{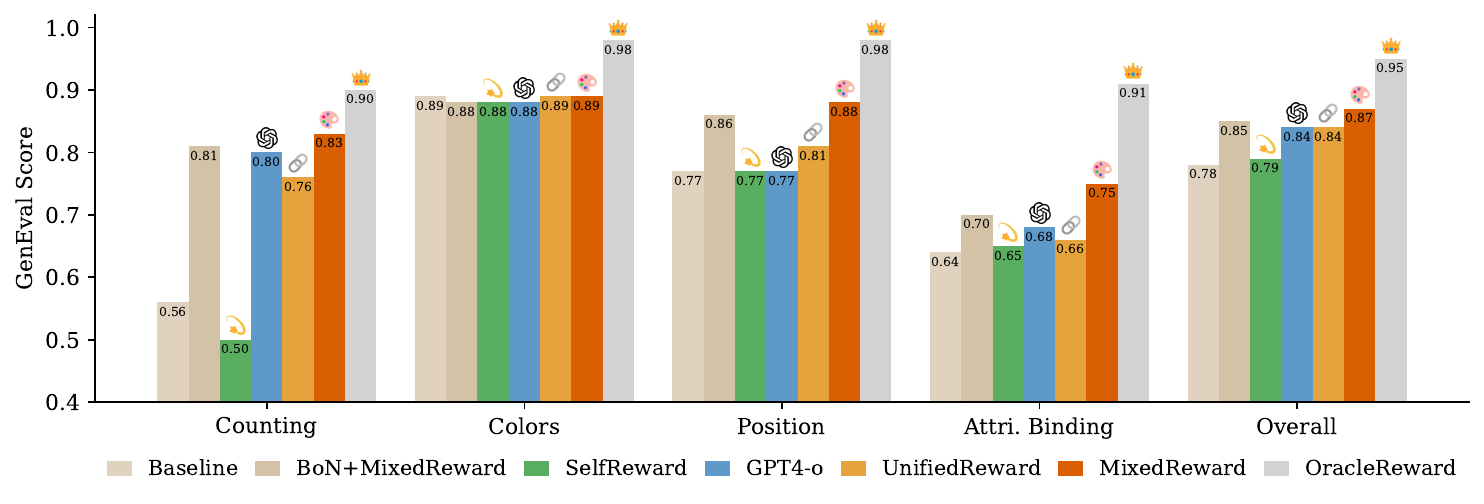}
  \vspace{-1.5em}
  \caption{Performance with different reward models on GenEval.}
  \label{fig:different_rewards_bar}
  \vspace{-1em}
\end{figure}

Unsurprisingly, \rewardgold gives rise to the best performance across all dimensions (see Figure~\ref{fig:different_rewards_bar}). For non-oracle critics, all variants surpass the baseline in terms of the overall score. Notably, MILR remains relatively robust to different reward models, except for \rewardself, which performs poorly on \textit{Counting} (around 0.5). Among non-oracle critics, \rewardmix performs the best, suggesting that, in the absence of oracle rewards, we can derive a strong universal reward model by combining specialized critic models. Moreover, \model\!\!+\rewardmix slightly outperforms the strong Best-of-N+\rewardmix baseline (+2.4\%) under comparable computation (i.e., N = T = 20), once again demonstrating the superiority of our method.

\subsubsection{Error Cases}
\label{sec:error_cases}

\begin{figure}[t!]
  \centering
  \includegraphics[width=1.0\linewidth]{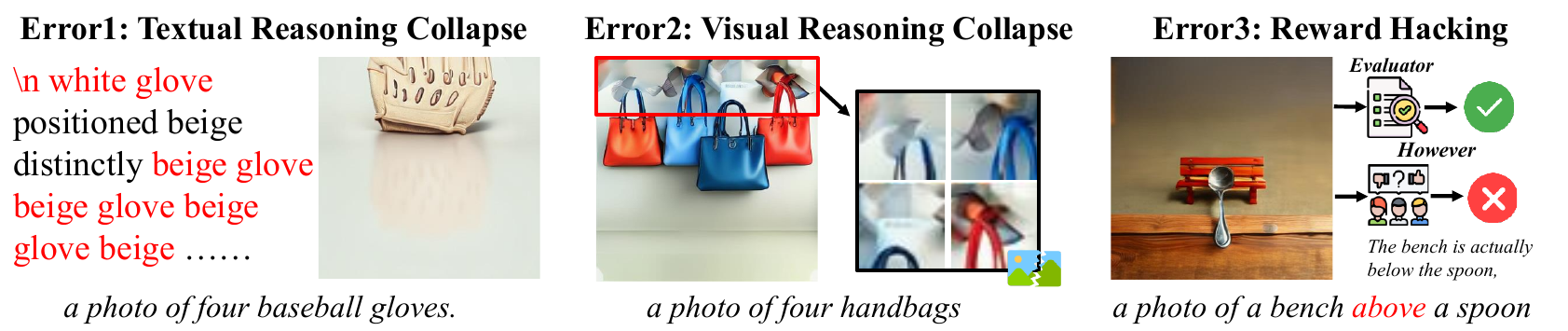}
  \vspace{-1.5em}
  \caption{Error Case Study}
  \label{fig:error_case}
  \vspace{-2em}
\end{figure}

Though \model has shown the best performance across all three benchmarks, we observe three major failure modes and illustrate them respectively in Figure~\ref{fig:error_case}. Specifically, the failure modes include
(1) \textbf{Textual Reasoning Collapse.} The regenerated chain of thoughts degenerates into repetitive phrases and becomes nonsensical for guiding image generation (e.g., repeated ``beige glove'').
(2) \textbf{Visual Reasoning Collapse.} While textual reasoning is fine, visual reasoning degrades prefix image tokens that govern the overall structure and subsequently adversely affects the generation of remaining tokens that control fine-grained details (e.g., blurred handles).   
(3) \textbf{Reward Hacking.} Models exploit shortcuts to achieve high rewards, but the generated image does not align with the given instruction perfectly (e.g., unmatched position relationship). This implies that the benchmark's evaluator is limited in that it is not yet fully capable of spatial reasoning.

% Despite its strong performance, we identify several distinct failure modes, as illustrated in Figure~\ref{fig:error_case}. The first category is Flawed Textual Reasoning, where the model's thought process degenerates into nonsensical or repetitive phrases (e.g., "beige glove beige glove..."), leading to an incoherent plan for image generation. A second error type is Visual Reasoning Collapse. In these cases, while the textual reasoning is coherent, the optimization process yields corrupted initial image tokens. This initial corruption then triggers a cascading failure during the subsequent autoregressive synthesis, degrading the entire image.
% Finally, we observe instances of Reward Hacking. Here, the model generates an image that successfully exploits the reward function to achieve a high score but fails to align with the intended semantics. For example, the model might incorrectly render the number of objects—such as generating two forks for a prompt requesting 'a fork'—yet still satisfy the benchmark evaluator. This type of failure highlights a significant limitation within the benchmark's own evaluation model, indicating a misalignment between its scoring criteria and the true semantic intent of the prompt.

\section{Discussion}
We acknowledge two primary limitations of \model. First, we have focused on an implementation of \model that builds upon autoregressive MUG, where both text and image tokens are generated autoregressively. Another strong paradigm of MUG is through diffusion image generation, where image tokens within the same image can attend to each other~\citep{deng2025bagel}. It is interesting to see if the effectiveness of \model transfers. Second, \model relies on a reward model for learning signals, but, in practice, a perfect reward model usually does not exist, and it is difficult to design a domain-agnostic reward model. In addition, our experimental results have revealed that the strongest non-oracle reward model still lags behind the oracle reward. Thus, future work is well-suited for designing reward models that can generalize like the unified reward model of~\citet{wang2025unifiedreward}.

% We acknowledge two primary limitations of the current work, which also suggest avenues for future research.
% First, the scope of our current method is focused on autoregressive MUG. While demonstrating strong results, its applicability to other architectures, such as the widely used diffusion models, has not yet been explored. 
% Second, the effectiveness of our method is inherently dependent on the quality of the reward signal, as current reward models adapted from task-specific evaluation tools often exhibit biases and may not generalize to broader requirements. Therefore, a critical area for future research is the development of more powerful and versatile reward models, such as self-improving systems where the understanding and generative components synergistically refine each other.

\section{Conclusion}
We have proposed \model, a test-time latent reasoning method for multimodal image generation. MILR employs policy gradient optimization, guided by an image quality metric, to search over the latent vector representations of discrete image and text tokens, leading to a unified multimodal reasoning framework. We implement \model with a unified multimodal understanding and generation model that natively supports language reasoning before image synthesis. Across three benchmarks, \model achieves state-of-the-art results. We further perform an in-depth analysis; we find that jointly reasoning in the latent image-text space is the key to its strong performance. Our qualitative studies also highlight \model's nontrivial capability in temporal and cultural reasoning. 

% approach that enhances compositional text-to-image generation by applying a policy gradient method at test time to jointly optimize shared text and image latent representations. Notably, MILR establishes a new state-of-the-art on GenEval and shows significant gains on T2I-CompBench and WISE, surpassing both costly fine-tuning methods and other test-time approaches. Through comprehensive ablation studies and analysis of key hyperparameters, our work validates test-time latent reasoning as a powerful and efficient paradigm for developing more capable generative models.


% \clearpage
\section{Ethics Statement}
% We adhere to the ICLR Code of Ethics. Our study uses only publicly available benchmarks and off-the-shelf models under their licenses; we do not construct new datasets, involve human subjects, or process personally identifiable information. All examples reported in the paper are synthetic.
We strictly adhere to the ICLR Code of Ethics. Below we elaborate on the ethical considerations relevant to our work and the measures we have put in place.
% our measures  followings are our practices regarding content safety and privacy/licensing.

\textbf{Safety of content.} To reduce potential risks, we confine our text prompts to those of the standard benchmarks, avoiding sensitive and offensive input. We manually reviewed the generated images and text used in the work and did not find harmful content.

\textbf{Privacy and intellectual property.} Our work does not involve processing personally identifiable information, as we experiment exclusively on publicly available benchmarks and off-the-shelf models, as per their respective licenses. 
% and we use all third-party assets and models in accordance with their respective licenses, with proper credit to prior work.

\section{Reproducibility statement}
To ensure reproducibility, we present experimental details, including settings and hyperparameter configurations, in~\Cref{sec:experimental_settings} and release our code at \url{https://github.com/spatigen/milr}.

% To ensure reproducibility, we provide detailed experimental settings and hyperparameter configurations in~\Cref{sec:experimental_settings}.

\bibliography{reference_header,iclr2026_conference}
\bibliographystyle{iclr2026_conference}

\clearpage
\appendix

%----------------------------------------------------------
%
%----------------------------------------------------------
\section{Appendix}
\subsection{Optimization algorithms}\label{appendix:algorithm}

By default, \model jointly optimizes text latent $\vz^{(t)}$ and image latent $\vz^{(v)}$ in each gradient descent step, which is denoted as \model-Joint. As an alternative, we can optimize $\vz^{(t)}$ and $\vz^{(v)}$ in an alternating fashion (denoted by \model-Alt), akin to the strategy used in coordinate descent.
% a strategy widely used for bilevel optimization problems. 
We further propose another variant of \model that first optimizes $\vz^{(t)}$ to its approximate optimum, followed by optimizing $\vz^{(v)}$ to its approximate optimum (denoted by \model-T2V). We illustrate the three algorithms in~\Cref{alg1}. Their performance on GenEval is presented in Table~\ref{benchmark:different_algorithms}. We do not see significant differences among them, so we use \model-Joint by default because of its simplicity.

\newcommand{\LO}{LatentReasoning}
\newcommand{\Joint}{\textsc{Joint}}
\newcommand{\Alternating}{\textsc{Alt}}
\newcommand{\TtoV}{\textsc{T2V}}

\begin{algorithm}
\caption{\emph{\model}}
\label{alg1}
\begin{algorithmic}
\Require Instruction $c$, Learning rate $\eta$, MUG model $p$, reward threshold $\tau$, text and image fraction $\lambda_{t},\lambda_{v} \in (0,1]$, optimization steps $K$, Reasoning strategy $\mathcal{S}\in\{\text{\Joint},\text{\Alternating},\text{\TtoV}\}$
\State $\vz, \vt,\vv \gets p(\vt,\vv|c)$ 
% \Comment{Initial latent vectors: \Cref{eq:inital}}
\Comment{Initial latent vectors}
\State $V_f\sim p(\cdot|\vv)$ 
\State $r\gets R(V_f, c)$ \Comment{Reward Calculation} 
\State $\vz^{0} \gets (z_{1:\lambda_{t} |\vt|}^{(t)};z_{1:\lambda_{v} |\vv|}^{(v)})$ \Comment{Set $\lambda_{t}$ and $\lambda_{v}$ fraction}
\State $k \gets 1$ \Comment{Starting step index}

\While{$k \leq K$ and $r \leq \tau$}
\State $V_f,\vz^{k} \gets \textcolor{red}{\LO}_{\mathcal{S}}\!\left(\vz^{k-1},\eta,k,K,c,p\right)$ \Comment{Select a strategy: \Cref{alg:schedules}}
\State $r \gets R(V_f, c)$
\State $k \gets k + 1$
\EndWhile

\State \Return $V_f$
\end{algorithmic}
\end{algorithm}

\newcommand{\AlgLegend}[1]{%
  \vspace{0.25em}%
  \noindent\begin{minipage}{\linewidth}
    \normalfont\normalsize\rmfamily % 和正文同字号/同字族
    \textbf{Note. }#1
  \end{minipage}%
  \vspace{0.25em}%
}

\begin{algorithm}
\caption{\LO\ (default: \Joint)}
\label{alg:schedules}

% \AlgLegend{$\mathcal{J}(\vz)$ as defined in ~\Cref{eq:milr-optimization}; $V_f$ sampled as in ~\Cref{eq:sampling_the_final_answer}; $T_f$ is the full reasoning text. \(\vz^{(t)}_k\) and \(\vz^{(v)}_k\) denote the text and visual latents at iteration \(k\).}

\begin{subalgorithm}[t]{.31\linewidth}
\caption{\Joint}
\begin{algorithmic}[1]\footnotesize
  \State $\vz^{k} \gets \vz^{k-1} + \eta \cdot \mathcal{J}(\vz^{k-1})$
  % \Comment{$\mathcal{J}(\vz^{k-1})$: \Cref{eq:milr-optimization}.}
  \State $V_f \sim p(V_f \mid \vz^{k}, c)$
  \State \Return $V_f,\vz^{k}$
\end{algorithmic}
\end{subalgorithm}\hfill
\begin{subalgorithm}[t]{.33\linewidth}
\caption{\Alternating}
\begin{algorithmic}[1]\footnotesize
  \If{$k \bmod 2 = 1$}
    \State $\vz^{(t)}_k \gets \vz^{(t)}_{k-1} + \eta \cdot \mathcal{J}(\vz^{(t)}_{k-1})$ 
    \State $T_f \sim p(T_f|\vz^{(t)}_k,c)$
    \State $V_f \sim p(V_f | \vz^{(v)}_{k-1},T_f, c)$
\Else
    \State $\vz^{(v)}_k \gets \vz^{(v)}_{k-1} + \eta \cdot \mathcal{J}(\vz^{(v)}_{k-1})$ 
    \State $V_f \sim p(V_f | \vz^{(v)}_k,T_f, c)$ 
\EndIf
\State \Return $V_f,\vz^{k}$
\end{algorithmic}
\end{subalgorithm}\hfill
\begin{subalgorithm}[t]{.33\linewidth}
\caption{\TtoV}
\begin{algorithmic}[1]\footnotesize
  \If{$k \le\lfloor K/2 \rfloor$}
      \State $\vz^{(t)}_k \gets \vz^{(t)}_{k-1} + \eta \cdot \mathcal{J}(\vz^{(t)}_{k-1})$
    \State $T_f \sim p(T_f|\vz^{(t)}_k,c)$
    \State $V_f \sim p(V_f | \vz^{(v)}_{k-1},T_f, c)$
    \Else
      \State $\vz^{(v)}_{k} \gets \vz^{(v)}_{k-1} + \eta \cdot \mathcal{J}(\vz^{(v)}_{k-1})$
      \State $V_f \sim p(V_f | \vz^{(v)}_k,T_f, c)$
    \EndIf
    \State \Return $V_f,\vz^{k}$
\end{algorithmic}
\end{subalgorithm}
\end{algorithm}

\begin{table}[H]
\centering
\caption{Geneval results with different reasoning strategies.}
\label{benchmark:different_algorithms}
\begin{adjustbox}{width=\linewidth}
\begin{tabular}
{l@{\hspace{0.5cm}}ccccccc}
\toprule
\multicolumn{1}{c}{\bf Algorithms} &
{ Single Obj.} &
{ Two Obj.} &
{ Counting } &
{ Colors } &
{ Position } &
{ Attr. Binding } &
{\bf Overall }
\\
\cmidrule{1-8}
\model-\Joint      & 1.00 & 0.96 & 0.90 & 0.98 & 0.98 & 0.91 & 0.95\\
\model-\Alternating & 1.00 & 0.98 & 0.87 & 0.95 & 0.97 & 0.88 & 0.94\\
\model-\TtoV    & 1.00 & 0.96 & 0.95 & 0.95 & 0.96 & 0.89 & 0.95\\
\bottomrule
\end{tabular}
\end{adjustbox}
\end{table}

\begin{table}[H]
\centering
\caption{Geneval results under random subset optimization of image tokens.}
\label{benchmark:random_image}
\begin{adjustbox}{width=\linewidth}
\begin{tabular}
{l@{\hspace{0.5cm}}ccccccc}
\toprule
\multicolumn{1}{c}{\bf Algorithms} &
{ Single Obj.} &
{ Two Obj.} &
{ Counting } &
{ Colors } &
{ Position } &
{ Attr. Binding } &
{\bf Overall }
\\
\cmidrule{1-8}
\model(Image+Prefix)      & 1.00 & 0.95 & 0.88 & 0.91 & 0.97 & 0.89 & 0.93\\
\model(Image+Random)      & 1.00 & 0.92 & 0.84 & 0.89 & 0.94 & 0.86 & 0.90\\
\bottomrule
\end{tabular}
\end{adjustbox}
\end{table}

\subsection{Qualitive Study}

\subsubsection{Reasoning trajectories}
\label{appendix:reasoning_traj}
\begin{figure}[h]
  \centering
  \includegraphics[width=1.0\linewidth]{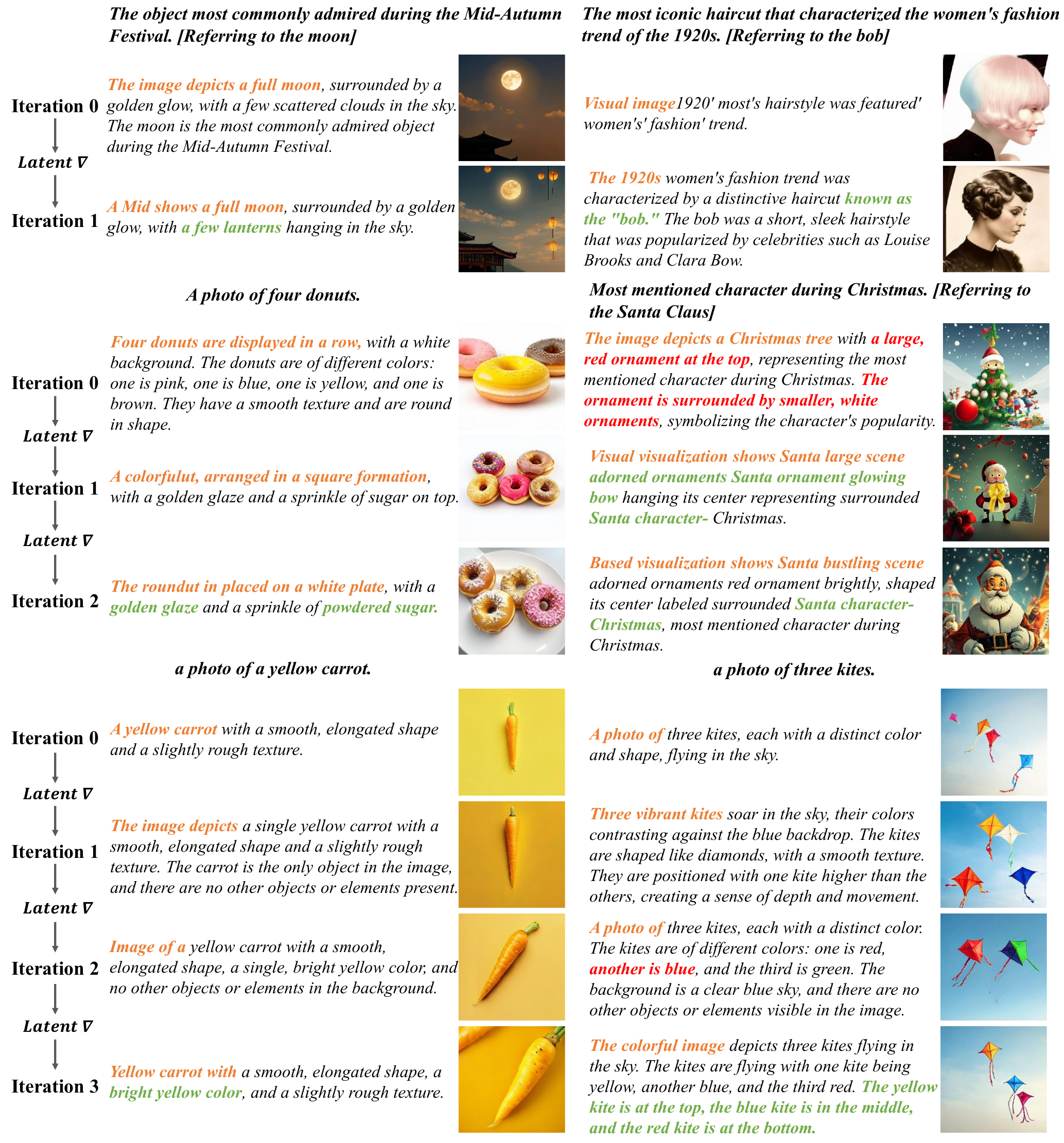}
  \caption{Example reasoning trajectories. Orange indicates prefix optimization, red highlights incorrect reasoning, and green represents correct reasoning.}
  \label{fig:case_study}
\end{figure}

\model admits multi-round latent-space refining at test time.
% , so some inputs require multiple iterations to reach the correct result.
In Figure~\ref{fig:case_study}, we visualize the full trajectories of rounds 2, 3, and 4 that finally arrive at the correct generation.
% two, three, and four rounds, qualitatively illustrating the gains across iterations.

\subsubsection{Model Ablations}
\begin{figure}[h]
  \centering
  \includegraphics[width=1.0\linewidth]{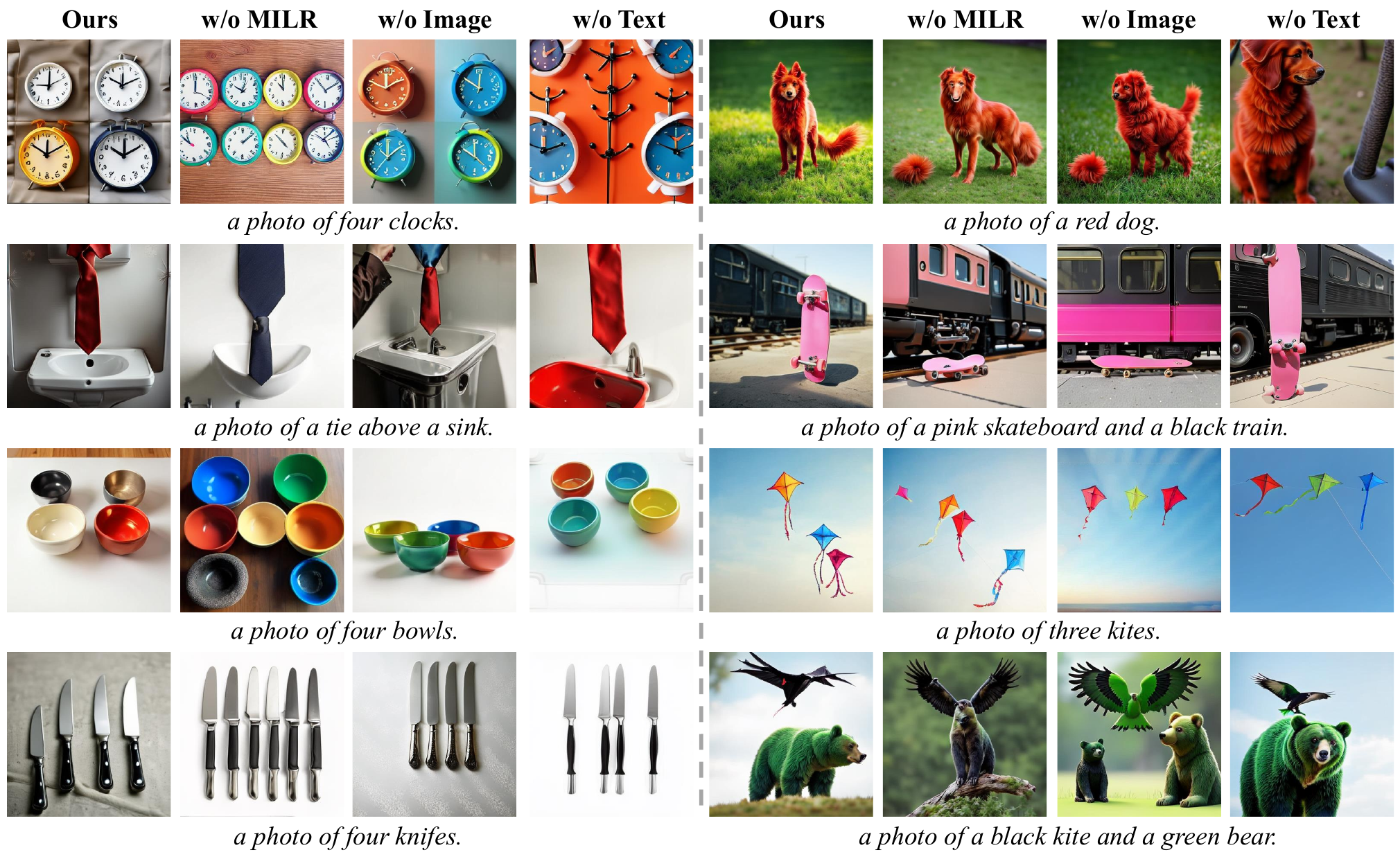}
  \caption{Example generations of \model (ours), the base model (w/o MILR), text-only (w/o Image) optimization, and image-only (w/o Text) optimization.}
  \label{fig:ablation_case_study}
    \vspace{-0.3em}
\end{figure}

We provide a qualitative study accompanying the model ablations done in~\Secref{sec:main-results}.
% We further provide more analysis and case study of ablation. Figure~\ref{fig:ablation_case_study} complements the main ablation in Table~\ref{ablations} by visualizing outputs for four settings.
Consistent with the quantitative results, single-modality optimization surpasses the base model but underperforms joint optimization.
Clearly, text-only optimization excels at tasks that rely on nontrivial numerical and compositional reasoning (e.g., “a photo of four clocks/bowls/knives” or “three kites”), while image-only optimization tends to refine image structures by adjusting the spatial arrangement of objects (e.g., “a photo of a tie above a sink”). To illustrate this scenario, consider the prompt ``a photo of \text{four clocks}" in Figure~\ref{fig:ablation_case_study}, text-only optimization produces four spatially uncorrelated clocks, whereas image-only optimization tends to piece together four visually similar clocks on the same wall. Interestingly, across all examples, image-only optimization exhibits a tendency to zoom in on the objects of interest.

% Figure~\ref{fig:ablation_case_study} complements the main ablation in Table~\ref{ablations} by visualizing outputs for four settings: \emph{Ours}, the base model (\emph{w/o MILR}), \emph{w/o Image} (text-only optimization), and \emph{w/o Text} (image-only optimization). Consistent with the quantitative results, both single-modality variants markedly surpass the base model, while optimizing both modalities yields the best outcomes. In prompts that require numerical or compositional reasoning (e.g., “a photo of four bowls/knives/vases” or “three kites”), \emph{w/o Image} often corrects counts and attributes yet may leave spatial relations imperfect; conversely, \emph{w/o Text} tends to refine layout (e.g., “a toilet left of a kite,” “a white boat and an orange hot dog”) but can drift in counts or colors. \emph{Ours} resolves both types of errors, illustrating the complementary roles of text- and image-side latent updates and explaining the overall gains when both are enabled.

\subsection{Additional analysis}

% \subsubsection{Performance breakdown on GenEval}
% We further break down model performance by category on GenEval. Clearly, text-only optimization excels at tasks that rely on nontrivial numerical and compositional reasoning (e.g., on \textit{counting} and \textit{color} tasks), while image-only optimization tends to refine image structures by adjusting the spatial arrangement of objects (e.g., on the \textit{position} task). To illustrate this scenario, consider the prompt ``a photo of \text{four clocks}" in Figure~\ref{fig:ablation_case_study}, text-only optimization produces four spatially uncorrelated clocks, whereas image-only optimization tends to piece together four visually similar clocks on the same wall. Interestingly, across all examples, image-only optimization exhibits a tendency to zoom in on the objects of interest. 

\subsubsection{Analysis of Text and Image Optimization Ratios}
\begin{figure}[ht]
  \centering
  \includegraphics[width=1.0\linewidth]{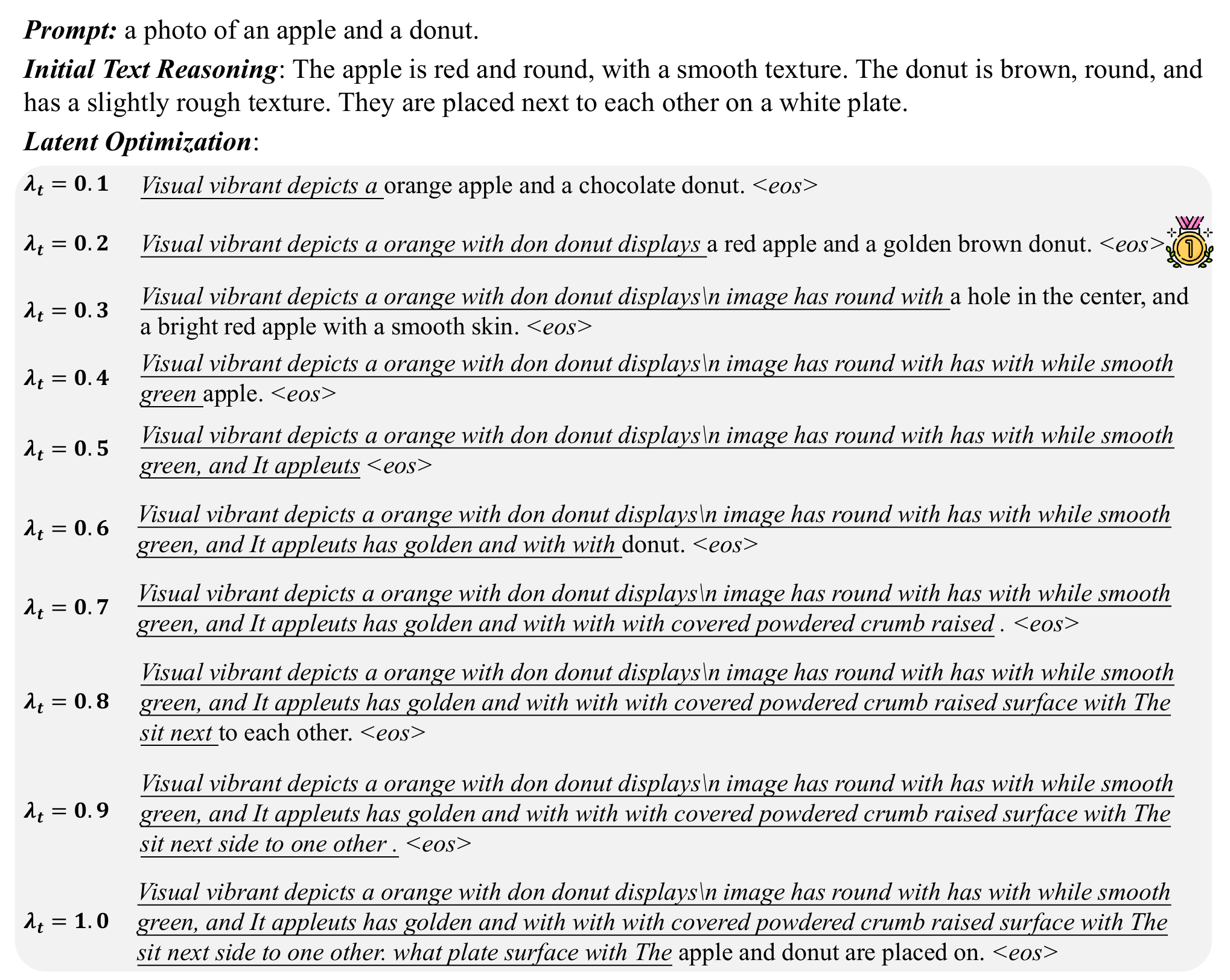}
  \caption{Case study of the text token optimization ratio $\lambda_t$. The underlined text is decoded from the optimized latents. \textit{\textless eos\textgreater} denotes the end-of-sentence token.
  % $\lambda_t=0.2$ gives the most reasonable result and is marked as the gold selection.
}
  \label{fig:text_cot_case}
\end{figure}

\begin{figure}[ht]
  \centering
  \includegraphics[width=1.0\linewidth]{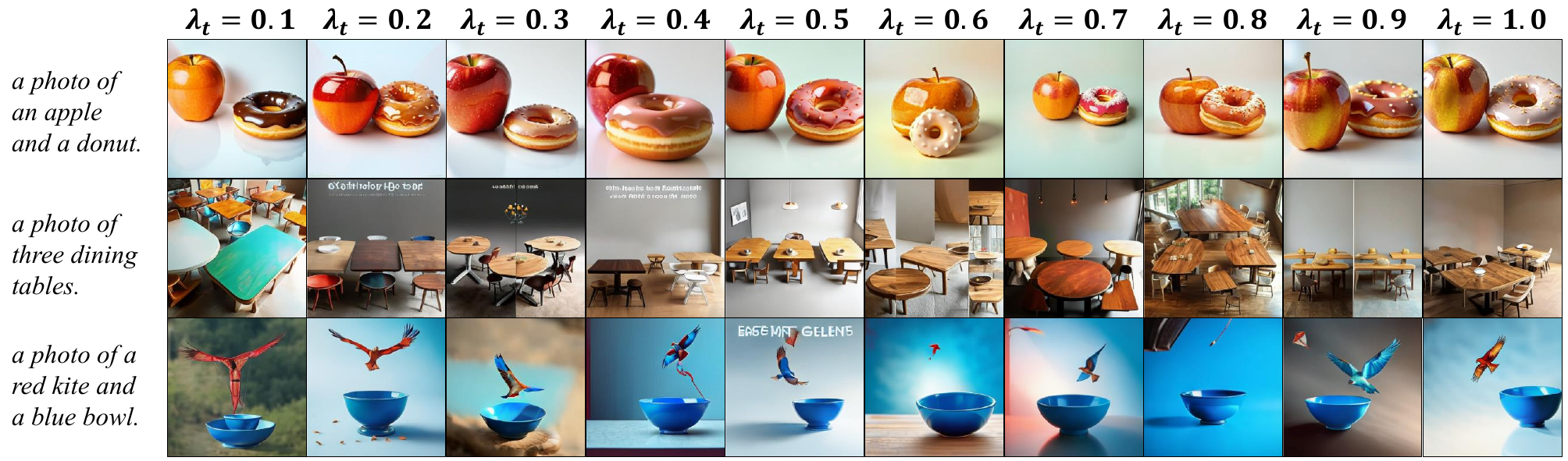}
  \caption{Generated images with varying token optimization ratio $\lambda_t$.}
  \label{fig:text_k_case}
\end{figure}

\begin{figure}[ht]
  \centering
  \includegraphics[width=1.0\linewidth]{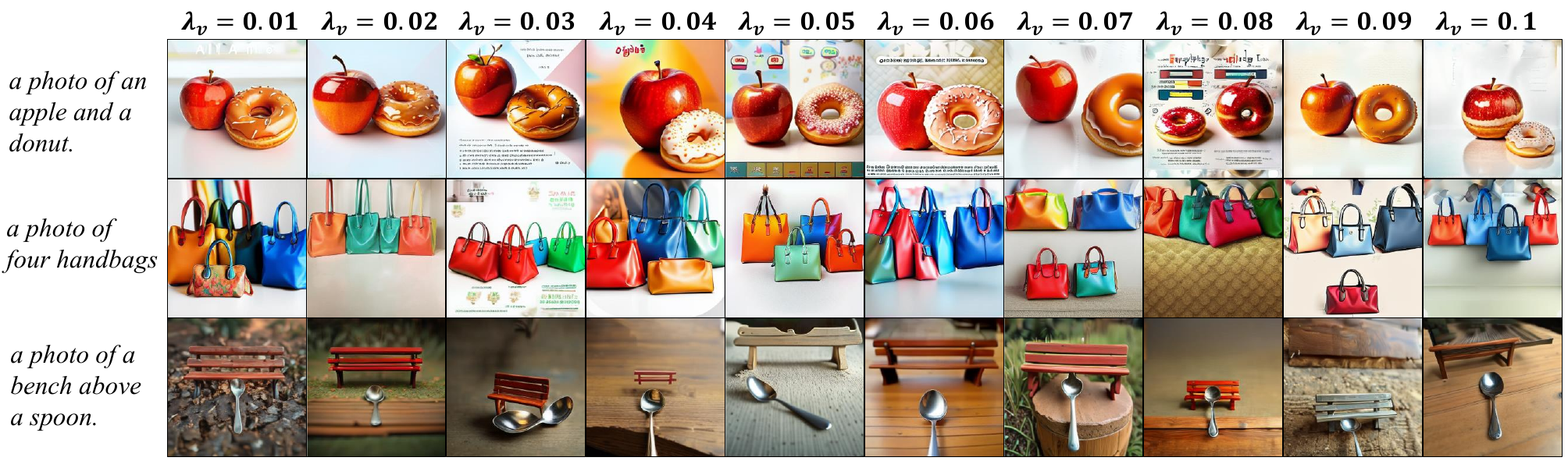}
  \caption{Generated images with varying image token optimization ratio $\lambda_v$.}
  \label{fig:image_k_case}
  
\end{figure}

In Section~\ref{sec:analysis-hyperparam}, we studied how the text-token ratio $\lambda_t$ and the image-token ratio $\lambda_v$ affect \model.
We also provide example language and image reasoning trajectories when varying $\lambda_t$ in Figure~\ref{fig:text_cot_case} and Figure~\ref{fig:text_k_case}, respectively.
In Figure~\ref{fig:image_k_case}, we show how image generation evolves when varying $\lambda_v$.
% The following figures complement the quantitative results with representative case studies.
% Figure~\ref{fig:text_cot_case} traces how the generated reasoning text evolves as $\lambda_t$ increases from 0.1 to 1.0. 
% Figure~\ref{fig:text_k_case} shows the corresponding visual outcomes across multiple prompts when varying $\lambda_t$. 
% Figure~\ref{fig:image_k_case} varies the image-token ratio $\lambda_v$ from 0.01 to 0.10. 
% These qualitative trajectories align with our quantitative findings (Figure~\ref{fig:text_image_analysis}): 
We empirically find that $\lambda_t=0.2$ and $\lambda_v=0.02$ generally lead to better generation.

\subsubsection{Analysis of Reward models}
\begin{figure}[h]
  \centering
  \includegraphics[width=0.8\linewidth]{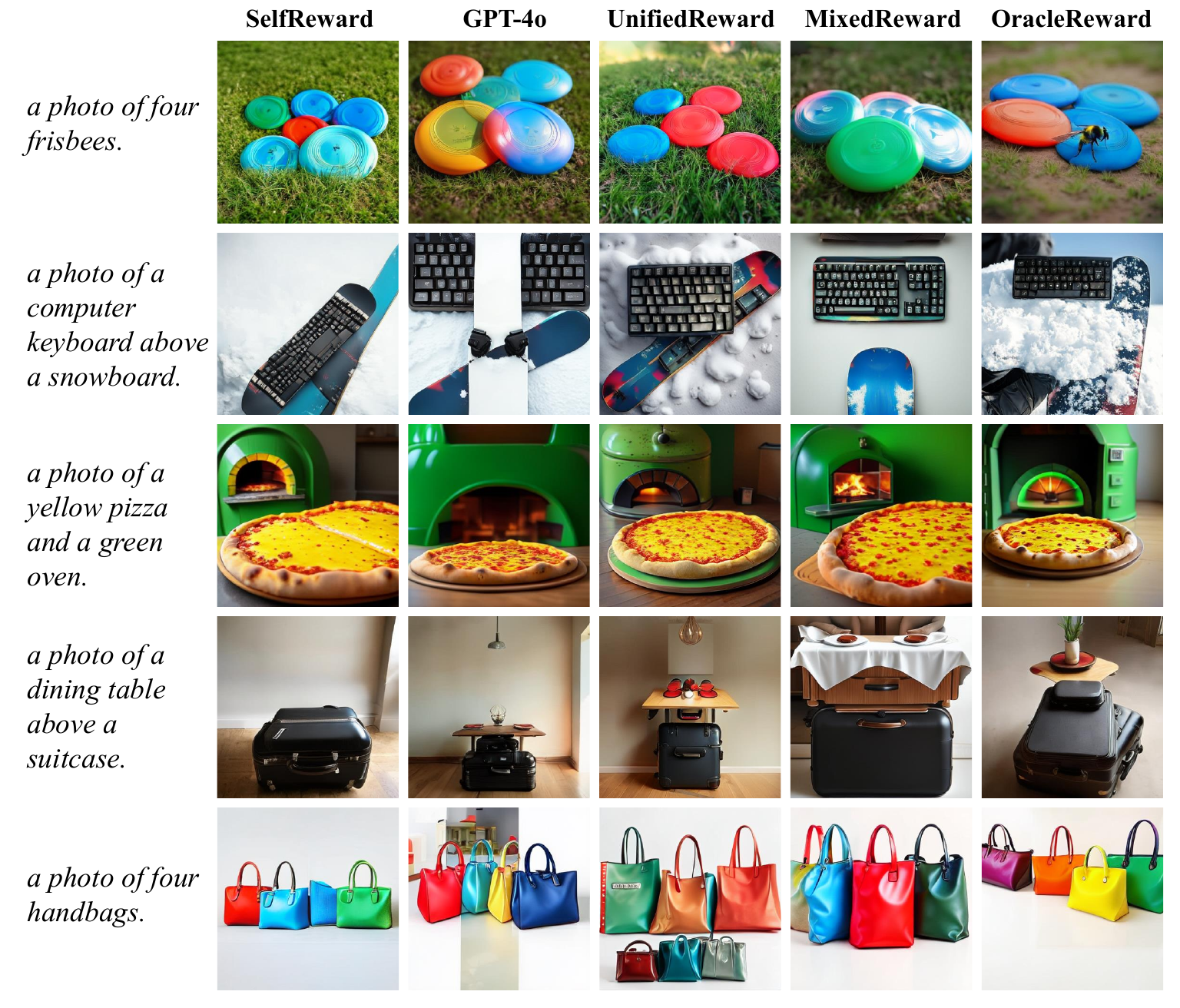}
  \caption{Case study of various reward models.}
  \label{fig:different_rewards}
\end{figure}

In Section~\ref{sec:reward-model}, we evaluated \model with different reward models: \rewardself SelfReward, \rewardtop GPT-4o, \rewardunify UnifiedReward, \rewardmix MixedReward, and \rewardgold OracleReward (see \Cref{benchmark:different_reward}, \Cref{benchmark:t2i-compbech} and \Cref{benchmark:wise}). We further provide example images generated under each setup in \Cref{fig:different_rewards}. Overall, \rewardgold establishes an upper bar of generation quality, and \rewardmix, among non-oracle reward models, performs best because it incorporates multiple specialized critics to provide a comprehensive assessment. Though \rewardmix still lags behind \rewardgold, in the absence of \rewardgold, we can compose a strong universal reward model from specialized off-the-shelf image critics. 

% This appendix complements Section~\ref{sec:reward-model} by evaluating \model under five reward models. \Cref{benchmark:different_reward} reports quantitative results across several task categories. Notably, the MixedReward achieves the best overall balance across tasks, whereas OracleReward provides an upper bound on performance.

% \Cref{fig:different_rewards} presents a qualitative comparison of reward functions, showing MixedReward as the most consistent across prompts. For the prompt ``a photo of four frisbees,'' SelfReward, GPT-4o, and UnifiedReward each miscount, whereas MixedReward correctly renders four; the OracleReward output even includes an unrelated bee. Overall, UnifiedReward typically surpasses SelfReward and GPT-4o in visual fidelity, supporting the value of evaluation-tuned rewards.

\begin{table}[H]
\centering
\caption{GenEval results with different rewards. The best performance is in bold.}
\label{benchmark:different_reward}
\begin{adjustbox}{width=\linewidth}
\begin{tabular}
{l@{\hspace{0.5cm}}ccccccc}
\toprule
\multicolumn{1}{c}{\bf Reward} &
{ Single Obj.} &
{ Two Obj.} &
{ Counting } &
{ Colors } &
{ Position } &
{ Attr. Binding } &
{\bf Overall }
\\
\cmidrule{1-8}
\hphantom{\rewardself}Baseline      & 0.98 & 0.85 & 0.56 & 0.89 & 0.77 & 0.64 & 0.78\\
\cmidrule{1-8}
\rewardself SelfReward      & \textbf{1.00} & 0.90 & 0.50 & 0.88 & 0.77 & 0.65 & 0.79\\
\rewardtop GPT-4o    & 0.98 & \textbf{0.94} & 0.80 & 0.88 & 0.77 & 0.68 & 0.84\\
\rewardunify UnifiedReward      & \textbf{1.00} & 0.92 & 0.76 & 0.89 & 0.81 & 0.66 & 0.84\\
\rewardmix MixedReward    & \textbf{1.00} & 0.90 & \textbf{0.83} & \textbf{0.89} & \textbf{0.88} & \textbf{0.75} & \textbf{0.87}\\
\cmidrule{1-8}
\rewardgold OracleReward    & 1.00 & 0.96 & 0.90 & 0.98 & 0.98 & 0.91 & 0.95\\
\bottomrule
\end{tabular}
\end{adjustbox}
\end{table}

\begin{table}[H]
\centering
\caption{T2I-CompBench results with different reward models. The best performance is in bold.}
\label{benchmark:t2i-compbech}
\begin{adjustbox}{width=\linewidth}

\begin{tabular}{l@{\hspace{0.5cm}}ccccccc}
\toprule
\multicolumn{1}{c}{\bf Reward} &
{ Color } &
{ Shape } &
{ Texture } &
{ Spatial } &
{ Non-Spatial } &
{ Complex } &
{\bf Overall}
\\
\cmidrule{1-8}
\hphantom{\rewardself}Baseline      & 0.6359 & 0.3528 & 0.4936 & 0.2061 & 0.3085 & 0.3559 & 0.3921\\
\cmidrule{1-8}
\rewardself Self Reward      & 0.7196 & 0.4620 & 0.5887 & 0.2379 & 0.3055 & 0.3868 & 0.4501 \\
\rewardtop GPT-4o           & 0.7624 & 0.4701 & 0.6561 & 0.2778 & \textbf{0.3102} & \textbf{0.3881} & 0.4775 \\
\rewardunify UnifiedReward    & 0.8208 & 0.4609 & 0.5835 & 0.3210 & 0.3044 & 0.3700 & 0.4651 \\
\rewardmix MixedReward      & 0.8009 & 0.4765 & 0.6608 & 0.4077 & 0.3055 & 0.3745 & 0.5043 \\
\cmidrule{1-8}
\rewardgold OracleReward    & \textbf{0.8508} & \textbf{0.5117} & \textbf{0.6949} &
                    \textbf{0.4613} & 0.3078 & 0.3684 & \textbf{0.5325} \\
\bottomrule
\end{tabular}

\end{adjustbox}
\end{table}

\begin{table}[H]
\centering
\caption{WISE results with different reward models. The best performance is in bold.}
\label{benchmark:wise}
\begin{adjustbox}{width=\linewidth}

\begin{tabular}{l@{\hspace{0.5cm}}ccccccc}
\toprule
\multicolumn{1}{c}{\bf Reward} &
{ Cultural } &
{ Time } &
{ Space } &
{ Biology } &
{ Physics } &
{ Chemistry } &
{\bf Overall}
\\
\cmidrule{1-8}
\hphantom{\rewardself}Base     
& 0.30 & 0.37 & 0.49 & 0.36 & 0.42 & 0.26 & 0.35 \\
\cmidrule{1-8}
\rewardself Self Reward      
& 0.40 & 0.43 & 0.53 & 0.38 & 0.44 & 0.23 & 0.41 \\
\rewardtop GPT-4o           
& 0.53 & 0.53 & 0.62 & 0.52 & 0.53 & 0.34 & 0.52 \\
\rewardunify UnifiedReward    
& 0.45 & 0.54 & 0.60 & 0.46 & 0.52 & 0.30 & 0.48 \\
\rewardmix MixedReward      
& 0.48 & 0.51 & 0.61 & 0.44 & 0.51 & 0.33 & 0.49 \\
\cmidrule{1-8}
\rewardgold Metric Reward    
& \textbf{0.64} & \textbf{0.65} & \textbf{0.72} &
  \textbf{0.66} & \textbf{0.71} & \textbf{0.37} &
  \textbf{0.63} \\
\bottomrule
\end{tabular}

\end{adjustbox}
\end{table}

\subsubsection{Efficieny Analysis}\label{app:efficiency}
We summarize the computing resources used by different image generation models in Table~\ref{tab:compute_resources}.
% summarizes the compute resources and training time required by different methods. 
Compared with training-based reasoning approaches such as T2I-R1~\citep{jiang2025t2i-r1}, Janus-Pro with DPO/GRPO~\citep{tong2025delving}, and Flow-GRPO~\citep{liu2025flowgrpo}, our test-time reasoning method achieves the best performance, without relying on curated reasoning data for training. As an example, Flow-GRPO requires about 2K A800 GPU hours for training but only matches \model, not to mention its labor cost in curating reasoning data. Moreover, when compared with the strong test-time reasoning method Best-of-N (N=20 vs. T=20 in \model), \model surpasses it by 0.04 while requiring less inference time because it employs an early-stop strategy, i.e., it stops once the generated image satisfies evaluation criteria.

\begin{table}[h]
\centering
\caption{Efficiency comparisons.}
\label{tab:compute_resources}
\resizebox{\linewidth}{!}{%
\begin{tabular}{lcccccc}
\toprule
\textbf{Method} & \textbf{\#GPU $\downarrow$} & \textbf{GPU} & \textbf{Training Time $\downarrow$} & \textbf{Inference Time $\downarrow$} & \textbf{Training Data} & \textbf{GenEval Score $\uparrow$} \\
\midrule
T2I-R1~\citep{jiang2025t2i-r1}         & 8  & H800  & 16 h        & --        & \cmark & 0.79 \\
Janus-Pro+GRPO~\citep{tong2025delving} & 8  & A100  & $\sim$9 h   & --        & \cmark & 0.81 \\
Janus-Pro+DPO~\citep{tong2025delving}  & 8  & A100  & $\sim$9 h   & --        & \cmark & 0.83 \\
Flow-GRPO~\citep{liu2025flowgrpo}      & 24 & A800  & $\sim$100 h & --        & \cmark & \textbf{0.95} \\
Reflect-DiT~\citep{li2025reflectdit}   & -- & A6000 & 24 h        & 16 h  & \cmark & 0.81 \\
Janus-Pro-7B(+Best-of-N, N=20)         & 1  & A100  & 0              & 8 h   & \xmark & 0.91 \\
\model~(\text{ours})                   & 1  & A100  & 0              & 5 h   & \xmark & \textbf{0.95} \\
\bottomrule
\end{tabular}%
}
\end{table}

\subsection{Latent Regularization to Mitigate Reward Hacking}
\label{app:latent-regularization}

As discussed in Section~\ref{sec:error_cases}, MILR can suffer from reward hacking, which refers to a scenario where a MUG model achieves high rewards but exhibits low image generation quality. To mitigate this issue, we adopt a simple $\ell_2$ regularizer to constrain the difference between the latents before and after optimization.
Let $\mathbf{z}_{\text{init}}$ denote the initial latents produced by a frozen MUG model before 
optimization, and let $\mathcal{J}(\mathbf{z}^k)$ denote the original objective used at the latents optimization step $k$ 
(see~\Cref{eq:milr-optimization}), we define the regularized objective as:
\begin{equation}
\mathcal{J}_{\text{reg}}(\mathbf{z}^k)
= \mathcal{J}(\mathbf{z}^k)
- \beta \big\|\mathbf{z}^k - \mathbf{z}_{\text{init}}\big\|_2^2\,,
\label{eq:latent-reg}
\end{equation}
where $\beta > 0$ controls the strength of regularization. Intuitively, this term encourages the optimized latents $\mathbf{z}^k$ to stay close to the initial latents, thus maintaining image generation capability similar to that of the frozen MUG model.
% reducing pathological updates that optimize for higher reward but lead to worse generation quality.

\begin{figure}[h]
  \centering
  \includegraphics[width=0.8\linewidth]{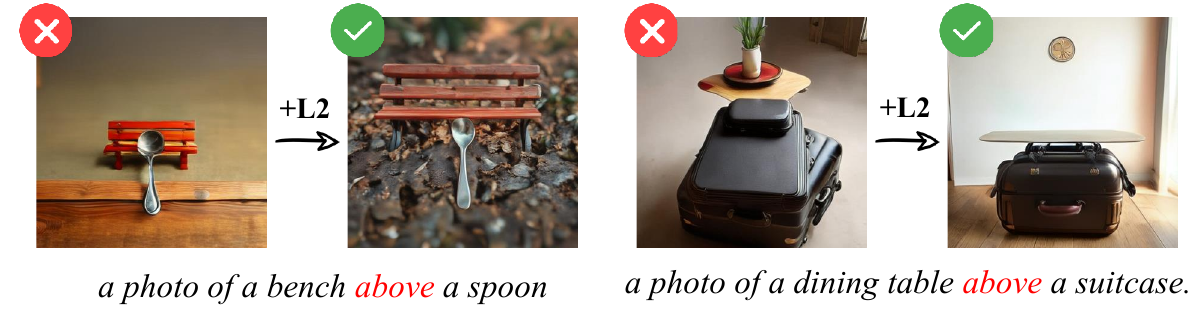}
  \caption{$\ell_2$ regularization on latents mitigates reward hacking.}
  \label{fig:reward_hacking}
\end{figure}

In our experiments, we use a small coefficient $\beta = 0.05$, which empirically stabilizes the optimization and alleviates semantic collapse, without degrading generation quality. 
% while leaving the main quantitative metrics essentially unchanged (0.94 in Overall). 
In \Cref{fig:reward_hacking}, we illustrate that adding latent regularization helps mitigate reward hacking. 
% This regularizer is simple to implement and does not modify the backbone parameters or the overall MILR pipeline.

\subsection{The Use of Large Language Models (LLMs)}
We used LLMs (e.g., OpenAI ChatGPT) only to polish our writing and to create \LaTeX{} algorithms and math equations. LLMs did not contribute to our research idea, model design, experimentation, etc. All contents of the paper have been verified by the authors.
% , who accept full responsibility.

\end{document}